\documentclass[letter, 10 pt, conference]{ieeeconf}  

\IEEEoverridecommandlockouts                              

\usepackage{amsmath}
\usepackage{algpseudocode}
\usepackage{algorithm}
\usepackage[dvips]{graphicx,color}
\usepackage{epsfig}
\usepackage{pstricks}
\usepackage{graphicx}
\usepackage{subfigure}
\usepackage{multicol}
\usepackage{url}
\usepackage{siunitx}
\usepackage{epstopdf}
\usepackage{booktabs}
\usepackage{multirow}
\usepackage{array}

\newsavebox{\ieeealgbox}

\title{\LARGE \bf An Improved Tabu Search Heuristic for Static Dial-A-Ride Problem}

\author{Songguang Ho, Sarat Chandra Nagavarapu, Ramesh Ramasamy Pandi and Justin Dauwels
\thanks{Songguang Ho is a research associate, Sarat Chandra Nagavarapu is a research fellow and Ramesh Ramasamy Pandi is a phd candidate with the School of Electrical and Electronic Engineering, Nanyang Technological University, 50 Nanyang Avenue, Singapore, 639798. {\tt\small nigelho@ntu.edu.sg, sarat@ntu.edu.sg,ramesh006@e.ntu.edu.sg}}%
\thanks{Justin Dauwels is an associate professor with the School of Electrical and Electronic Engineering, Nanyang Technological University, 50 Nanyang Avenue, Singapore, 639798. {\tt\small jdauwels@ntu.edu.sg}}%
\thanks{The research was partially supported by the ST Engineering $-$ NTU Corporate Lab through the National Research Foundation (NRF) corporate lab@university scheme.}
}
\IEEEoverridecommandlockouts
\begin{document}

\maketitle
\thispagestyle{empty}
\pagestyle{empty}


\begin{abstract}
Multi-vehicle routing has become increasingly important with the rapid development in autonomous vehicle technology. Dial-a-ride problem, a variant of vehicle routing problem (VRP), deals with the allocation of customer requests to vehicles, scheduling the pick-up and drop-off times and the sequence of serving those requests by ensuring high customer satisfaction with minimized travel cost. In this paper, we propose an improved tabu search (ITS) heuristic for static dial-a-ride problem (DARP) with the objective of obtaining high quality solutions in short time. Two new techniques, construction heuristic and time window adjustment are proposed to achieve faster convergence to global optimum. Various numerical experiments are conducted for the proposed solution methodology using DARP test instances from the literature and the convergence speed up is validated.

\end{abstract}

\section{Introduction}
\label{sec:intro}
Dial-A-Ride Problem (DARP) addresses the issue of door-to-door transportation service for the customers with high customer satisfaction. Now-a-days, transportation services have increasing need in our daily life, and it started to directly impact our environment as well as quality of living. According to a study conducted by University of British Columbia, the road pricing or pay-per-use is the most effective way to reduce emissions and traffic \cite{1}. DARP has many applications ranging from taxi services to autonomous cargo and ground operations at the airports.

DARP is an extension of pick-up and delivery problem under the class of vehicle routing problem (VRP) \cite{15}. It is a combinatorial optimization problem with an objective function to minimise the overall cost while satisfying a specific set of constraints such as time-window, maximum waiting time and maximum ride time to ensure high-quality customer service. In this problem, a set of customers makes a request for pick-up and drop-off at certain locations within a predefined time-window. An approach to solve DARP based on dynamic programming has been proposed in \cite{3}, in which divide and conquer method is used to solve the problem. However, the exponential relation between computational complexity and instance size severely limits the applicability to small scale instances.

Branch and cut is a classical way of solving mathematical optimization problems. A new branch-and-cut method \cite{4} was proposed which employs cutting planes to produce mathematical formulation for DARP. The problem was initially formulated using Mixed Integer Linear Programming (MILP). Later Cordeau et. al. \cite{2} have revisited the mathematical formulation for DARP and the research community has addressed this version as standard DARP. A modified approach based on branch-and-cut algorithm \cite{5} for pick-up and delivery problems with time windows was solved for larger instances to achieve optimality. Recently, another variant of branch-and-cut \cite{6} with less compact modeling helped to solve the previously unsolved benchmark instances for the heterogeneous-DARP to optimality within a matter of seconds.

Though the exact methods are useful to optimally solve the problem, the computational complexity of such methods is very high. Therefore, heuristic and meta-heuristic approaches have become widely used techniques to solve DARP. One such technique proposed in \cite{7}, which has pioneered the heuristic approaches. A sequential insertion algorithm \cite{8} has been designed for static DARP, which analyses the problem complexity, while offering flexibility to users. The literature also has archives of parallel insertion heuristics using the distributed computing technologies \cite{14}.

Dynamic fuzzy logic, a computationally efficient heuristic method was adopted to solve DARP \cite{9}. The tabu search heuristic \cite{2} aims to progressively explore the neighborhood structure from an initial solution while forbidding some solutions with similar attributes of recently visited solutions. It can be described as cleverly guided local search that is efficient in exploring the search space at the hope of finding global optimal solutions. However, the main drawback is the time spent while getting stuck in sub-optimal solutions. In order to avoid that, a parallel tabu search heuristic was proposed in \cite{11}. When infinite penalty is considered, the best solution is obtained irrespective of irrelevance in choice of initial static solution for the dynamic dial a ride problem.

A two phase heuristics approach \cite{12} based on insertion and improvement phase was proposed for DARP by Beaudry et al. From the solution obtained by insertion heuristics, the algorithms selects best non-tabu solution progressively while performing both inter and intra-route neighborhood evaluations. Recently, another variation of tabu search, named as granular tabu search algorithm \cite{13} that produces good solutions in short amount of time within 2-3 mins has been introduced. Also, this new method produces better results when compared to the classical tabu search, genetic algorithm and variable neighborhood search techniques.

In this paper, we propose an improved tabu search (ITS) heuristic for dial-a-ride problem. We assume that the routing time and cost from each vertex to every other vertex are known apriori. The major contributions of the paper are summarized as follows:
\begin{itemize}
\item Determining a computationally faster and reliable variant of tabu search for DARP by thorough investigation of various neighborhood evaluation and insertion techniques.
\item Proposed a new construction heuristic for tabu search to obtain good initial solution rapidly.
\item Designed a time window adjustment technique for faster solution convergence.
\item Implemented and tested the proposed techniques using various DARP test instances to verify the acceleration in convergence.
\end{itemize}

The remainder of the paper is organized as follows: Section~\ref{sec:formulation} briefly discusses the DARP mathematical formulation. Section~\ref{sec:tabu} details tabu search heuristic and several variants of tabu search using neighborhood evaluation and insertion techniques. Section~\ref{sec:modified} presents the proposed ITS heuristic method. Section~\ref{sec:simulation} illustrates the convergence analysis results for the proposed algorithm. Conclusions are provided in Section~\ref{sec:conclusion}.

\section{Problem Formulation for DARP}
\label{sec:formulation}
Dial-a-ride problem (DARP) is a variant of VRP that involves dispatching of a fleet of vehicles to transport customers between their desired pick up and drop off locations within specified time windows. The aim is to minimise the over-all transportation cost of the vehicle by evading the longest route in tandem with providing superior passenger comfort and safety. DARP is mathematically formulated as an optimization problem with an objective function subjected to several constraints.

In dial-a-ride problem, $n$ customer requests are served using $m$ vehicles. Each request $i$ consists of time window either for departure or arrival vertex. The objective is to minimise the travel cost subject to several constraints. Let $S$=$\{s_1$,$s_2$,$\cdots\}$ denotes the solution space. All solutions $s_i$ $\in$ $S$ need to satisfy three basic constraints. Every route for a vehicle $k$ starts and ends at the depot and the departure vertex $v_i$ and arrival vertex $v_{i+n}$ must belong to the same route, and the arrival vertex $v_{i+n}$ is visited after departure vertex $v_i$. Any solution that violate these basic set of constraints becomes infeasible. In addition, several other constraints need to be satisfied; the load of vehicle $k$ cannot exceed preset load bound $Q_k$ at any time; the total route duration of a vehicle $k$ cannot exceed preset duration bound $T_k$; the ride time of any passenger cannot exceed the ride time bound $L$; the time window set by the customer must not be violated.

Four major constraints exist in dial-a-ride problem: load, duration, time window and ride time constraints. Load constraint violation $q(s)$ occurs when the number of passenger in a vehicle $k$ exceeds its load limit $Q_k$; duration constraint violation $d(s)$ happens when a vehicle $k$ exceeds its duration limit $T_k$; time window constraint violation $w(s)$ appears when the time constraint is violated; ride time constraint violation $t(s)$ occurs when a passenger is transported for a longer time than ride time limit $L$. The constraints are given by Eq. (1-4).
\begin{align}
q(s) &= \sum_{\forall k} \max(q_{k,max}-Q_k,0), \\
d(s) &= \sum_{\forall k} \max(d_k-T_k,0), \\
w(s) &= \sum_{\forall i} \big[\max(B_i-l_i,0) + \max(B_{i+n}-l_{i+n},0)\big], \\
t(s) &= \sum_{\forall i} \max(L_i-L,0).
\label{eq:constraints}
\end{align}

The next section presents the variants of tabu search based on the neighborhood evaluation and insertion techniques.
\section{Tabu Search Heuristic}
\label{sec:tabu}

Tabu search is a higher level heuristic procedure to solve optimization problems, as originally defined by \cite{10}. The methodology is efficient at escaping local optimal solutions with structured memories and tabu list. During the neighborhood transitions, cycling should be avoided to intelligently explore the search space for global optimal solutions. Intensification and diversification strategies have to be properly employed in order to attain optimal solutions. In tabu search, the recent neighborhood transition is recorded in a tabu list. On subsequent transitions, the recorded moves in tabu list is not considered. In this way, cycling of moves gets avoided. However, for some cases, if the objective function is below the best obtained cost, then aspiration level performs the move even if it is in tabu list.

\cite{2} has successfully superimposed the methodology on local search with neighborhood reduction to solve DARP. In this work, the objective function is considered as travel cost with weighted penalties for additional constraints. At this case, the method could efficiently explore the search space through some infeasible solutions in the hope of finding global optimal solutions.
\begin{align}
f(s) &= c(s) + \alpha q(s) + \beta d(s) + \gamma w(s) + \tau t(s).
\label{eq:objective}
\end{align}

The objective function $f(s)$ is given by Eq. 1, where $c(s)$ is the travel cost and $\alpha$, $\beta$, $\gamma$, $\tau$ are the penalty coefficients which are initialized to 1. These penalty coefficients change periodically once the optimization process begins. In tabu search, the moves that result in both feasible and infeasible solutions are accepted during the optimization process. When a solution violates the constraints and become infeasible, the penalty coefficient for those constraints are increased by a factor $(1+\delta)$ and decreased by the same factor, when the constraint is not violated. In this way, the algorithm intelligently explores the search space in the direction where the constraints that are violated during the previous moves are relaxed. 

In this paper, construction of initial solution, route optimization, intensification, diversification and the evaluation functions are adopted from Cordeau et al. \cite{2}.

\subsection{Neighborhood Evaluation}
\label{sec:level}
\begin{figure}[h]
	\centering
	\includegraphics[scale=0.54]{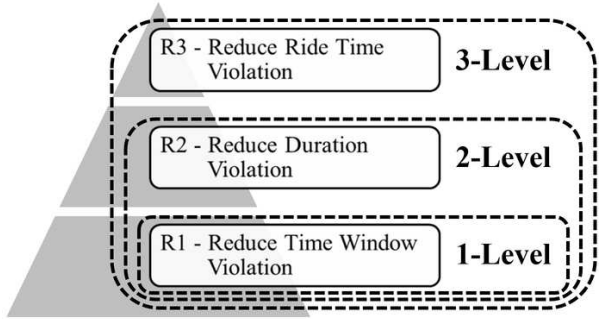}
    \caption{Neighborhood evaluation for tabu search.}
    \label{fig:neighborhood}
\end{figure}

An important aspect of Tabu search heuristic is the procedure for neighborhood evaluation technique. The objective of neighborhood evaluation is to reduce the constraint violations and assess the feasibility of the solution. There are three objectives in neighborhood evaluation: Reduction of time-window constraint associated with requests (R1); Reduction of route duration constraint associated with vehicle (R2); Reduction of Ride Time Constraint associated with requests (R3). Fig. \ref{fig:neighborhood} depicts the three level neighborhood evaluation method for tabu search.

There are three neighborhood evaluation techniques: i) 1-level ii) 2-level and iii) 3-level, which are `steps(1-2)', `steps(1-6)' and `full procedure' adopted from \cite{2} in their respective order. The objectives for each level is substrate-based i.e., each level is built upon subsequent levels. 1-level has the objective R1, 2-Level has the objective R1 $\&$ R2 and 3-Level has objective R1, R2 $\&$ R3.

\subsection{Insertion Techniques}
\label{sec:insertion}
\begin{figure}[H]
	\centering
	\includegraphics[scale=0.57]{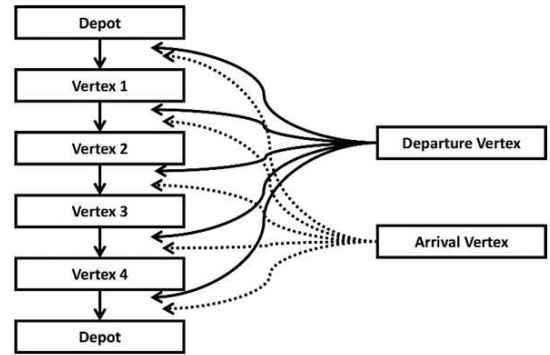}
    \caption{Insertion technique.}
    \label{fig:one_step}
\end{figure}

The size of neighborhood directly influence the computational complexity required during the optimization process. There are two insertion techniques carried out in this paper: i) one-step insertion and ii) two-step insertion. The convention was adopted due to nature of the neighborhood transition, which are: a) Single paired insertion (SPI) \cite{16} and b) Neighborhood Reduction \cite{2}. Fig. \ref{fig:one_step} illustrates the insertion of departure and arrival vertices into the routes.

For one-step insertion, the algorithm attempts to move each vertex pair from one vehicle to another in a single transition. After insertion of the vertex pair, the neighborhood solutions are evaluated using the objective function in Eq. 5. The neighbor with minimum cost is selected as the next move. Though the search process is very expensive, this type of move has the greatest potential for improvement in the objective function \cite{16}.

For two-step insertion, the algorithm attempts to move each vertex pair from one vehicle to another in two sequential steps. For each request, critical vertex indicates pick-up or drop-off point that consists of narrower time window when compared to the non-critical vertex. The first step is the insertion of critical vertex into its best position, which gives with least value of objective function in Eq. 5. While holding the current position, the second step is the insertion of non-critical vertex into its best position. The second step consists of two possibilities: i) if critical vertex is departure node, then insert non-critical vertex only after the critical vertex, ii) if critical vertex is arrival node, then insert non-critical vertex only before the critical node. This technique significantly reduce the neighborhood from moves $O(r^2)$ to $O(r)$, where $r$ is number of vertices in route $k$.


\begin{table}[h]
\centering
{
\begin{tabular}{>{}c*{3}{c}}
\toprule
 & One Step Insertion & Two Step Insertion \\
\midrule
1-Level Evaluation & $\text {TS}_{11}$ & $\text {TS}_{12}$ \\
2-Level Evaluation & $\text {TS}_{21}$ & $\text {TS}_{22}$ \\
3-Level Evaluation & $\text {TS}_{31}$ & $\text {TS}_{32}$ \\
\bottomrule
\end{tabular}}
\caption{Comparison of the median of travel cost: Tabu Search (TS) \cite{2} vs Improved Tabu Search (ITS).}
\label{tab:TS_Variants}
\end{table}
Based on the mentioned neighborhood evaluation and insertion techniques, the possible combinations are listed in Table~\ref{tab:TS_Variants}. Each method is represented using the convention $\text{TS}_\text{NI}$, where $N$ represents the neighborhood evaluation level and $I$ represents the insertion step.

\begin{figure}[H]
	\centering
	\includegraphics[scale=0.18]{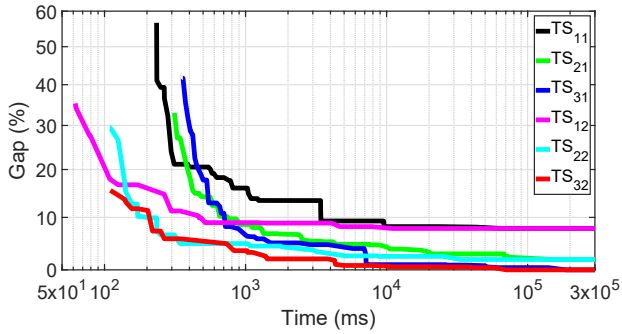}
    \caption{Convergence Analysis for TS using R1a.}
    \label{fig:R1a_1_TS}
\end{figure}
\begin{figure}[h]
	\centering
	\includegraphics[scale=0.18]{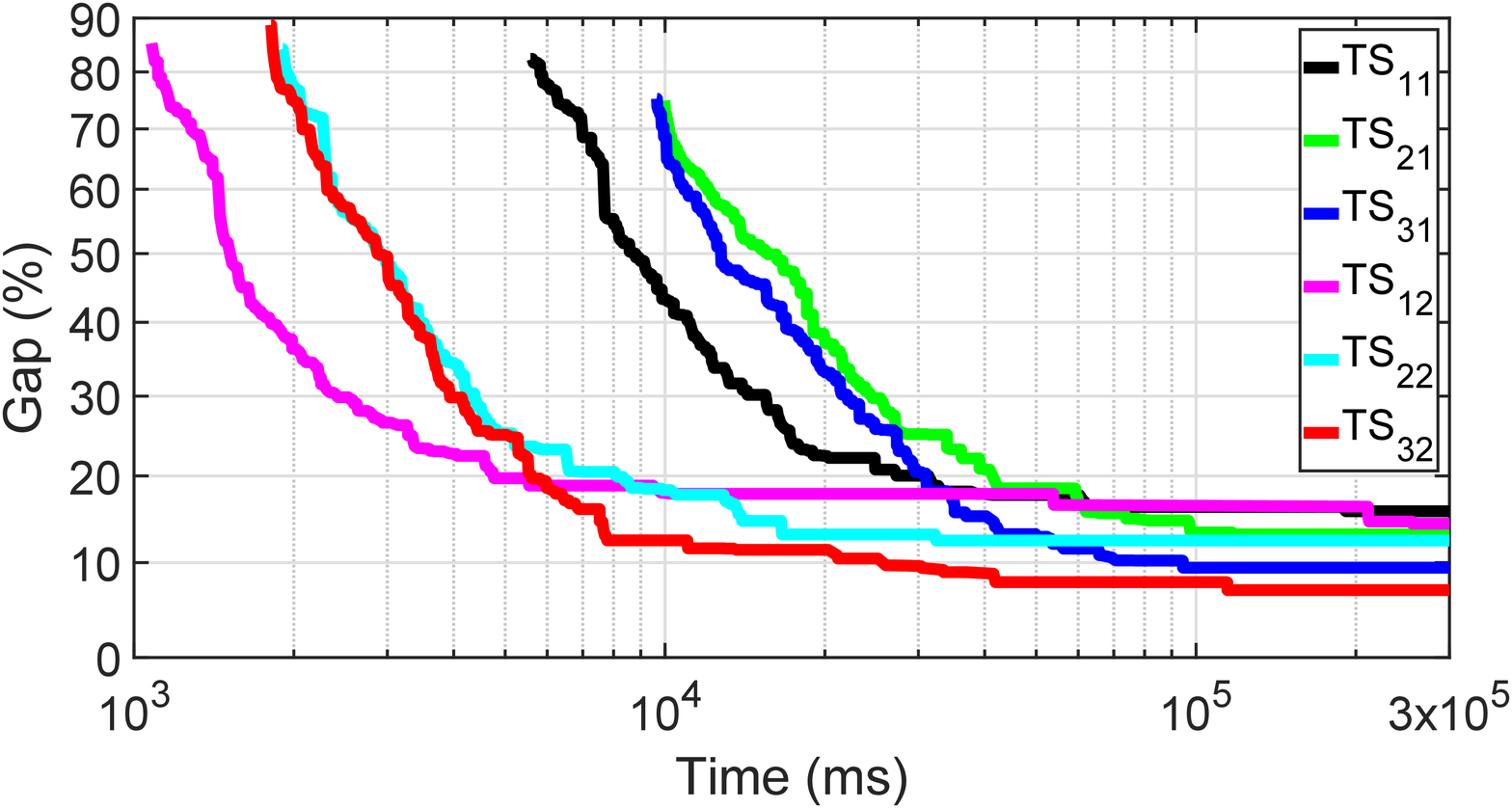}
    \caption{Convergence Analysis for TS using R3a.}
    \label{fig:R3a_1_TS}
\end{figure}
\begin{figure}[h]
	\centering
	\includegraphics[scale=0.18]{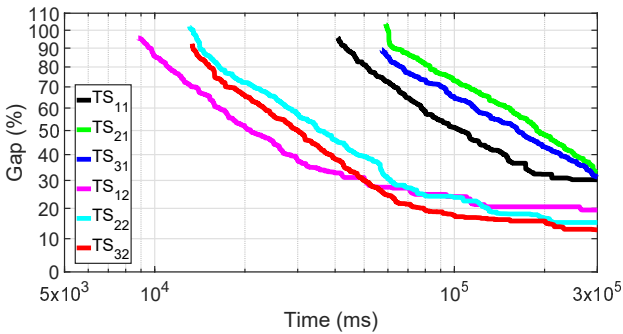}
    \caption{Convergence Analysis for TS using R6a.}
    \label{fig:R6a_1_TS}
\end{figure}

Various benchmark instances for standard DARP are provided by Cordeau and Laporte in \cite{2}. In these instances, the number of requests vary between 24 to 144 and the fleet size varies between 3 to 13. The capacity of each vehicle is set to 6, the maximum passenger ride time is 90 min, the maximum route duration is 480 min and the route planning horizon is 24 hrs. 

The costs of the benchmark solutions provided by Parragh and Schmid \cite{17} for the instances R1a, R3a and R6a are 190.02 , 532.00 and 785.26 respectively. Figs. \ref{fig:R1a_1_TS}, \ref{fig:R3a_1_TS}, \ref{fig:R6a_1_TS} illustrate the convergence of each of these tabu search variants validated using these instances. Each colored line in the plot corresponds to a distinct variant of tabu search as listed in Table \ref{tab:TS_Variants}. The x-axis of the plots corresponds to the simulation run time and the y-axis represents the gap (\%) as given by (\ref{eq:cost_difference}). We plot the median of the gap recorded from five independent simulations ran for a duration of five minutes. Where, `BKS' corresponds to the best known solution for the problems. We decided to consider the median in order to restrict the effect of outliers on the analysis.

\begin{align}
\text{Gap (\%)}~=~\frac{\text{cost}~-~\text{BKS}}{\text{BKS}}\times100.
\label{eq:cost_difference}
\end{align}


From the convergence plots, the following inferences are observed:
1) Two-step insertion technique has faster convergence than One-step insertion.
2) One-level achieves initial solution in the least time, followed by Three-level and Two-level neighborhood evaluation techniques.
3) Three-level neighborhood evaluation technique converges deeper over time when compared to other techniques, which is followed by Two-level, while One-level is the slowest.

We consider that $\text{TS}_\text{32}$ as better variant as it converges to optimality in a lesser time, while employing three-level neighborhood evaluation and two-step insertion techniques. However, $\text{TS}_\text{32}$ takes more time to obtain a feasible initial solution. In order to address this issue, new techniques are proposed in the next section.

\section{Improved Tabu Search (ITS)}
\label{sec:modified}
Tabu search heuristic has been extensively used in dial-a-ride problem due to its comparatively faster execution. However, the time required is still significant, especially for larger instances. So, there is a need to optimize the algorithm in order to obtain good results in a reasonable time. From the analysis presented in Section \ref{sec:tabu}, it is clear that the algorithm takes longer time to converge and to obtain a feasible initial solution. In order to overcome this issue, two new methodologies are proposed: a) construction heuristic (CH) and b) time window adjustment (TW). The main objective of the improved tabu search is to obtain high quality solutions within a short time. The Sections \ref{sec:construction} and \ref{sec:TW} detail the proposed methodologies.

\subsection{Construction Heuristic}
\label{sec:construction}
The convergence analysis presented for tabu search in Section \ref{sec:tabu} indicates that the time taken to find the first feasible solution for a problem highly depends on the quality of initial solution. In \cite{2}, a random initial solution is generated to start the search for global optimal solution. However, such methodology has higher tendency to random seed, and it takes longer time to find first feasible solution.

In this paper, a new construction heuristic is proposed to find high quality solutions more rapidly. In this method, an empty set of routes is created, and requests are sorted randomly. After the initial preparation, each request is inserted sequentially into the position that attains minimum objective function. Therefore, the objective function is formulated as $f(s)$ in \ref{eq:objective}, with $\alpha$ = $\beta$ = $\gamma$ = $\tau$ = 1. The details of the proposed construction heuristic technique are described using Algorithm \ref{al:CH}.

\begin{algorithm}
\caption{\textit{Construction Heuristic (CH)}}
\label{al:CH}
\begin{algorithmic}[1]
\Require Number of request ($n$), number of vehicle ($m$), all constraints
\Ensure Initial solution for ITS
\State Parameter initialization: set $\alpha$ = $\beta$ = $\gamma$ = $\tau$ = 1.
\State \textit{random\_list} = sort requests in random order.
\State Initialize the vehicles with empty set of routes.
\For {all requests in \textit{random\_list}}
    \For {all vehicles}
        \State Try inserting request in all possible positions.
    \EndFor
    \State Select an insertion with least $f(s)$.
    \State Update the solution.
\EndFor			
\end{algorithmic}
\end{algorithm}

The next section discusses the proposed time window adjustment.

\subsection{Time Window Adjustment}
\label{sec:TW}
According to the benchmark instances from \cite{2}, the DARP problem is modeled as set of requests with time-window constraints for optimal allocation to vehicles. The objective is to minimise the travel cost, while satisfying the constraints such as time window and ride time constraints associated with requests, and route duration constraint associated with vehicles.

\begin{figure}[h]
	\centering
	\includegraphics[scale=0.54]{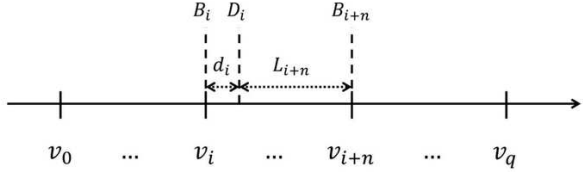}
    \caption{Ride time $L_i$ for request $i$.}
    \label{fig:TW_Ridetime}
\end{figure}

\begin{figure}[H]
	\centering
	\includegraphics[scale=0.54]{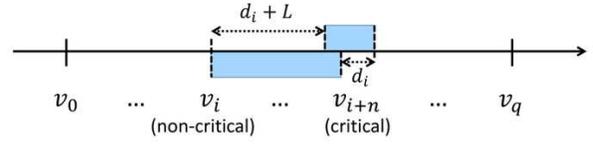}
    \caption{Time window adjustment for departure vertex.}
    \label{fig:TW_Departure}
\end{figure}

\begin{figure}[H]
	\centering
	\includegraphics[scale=0.54]{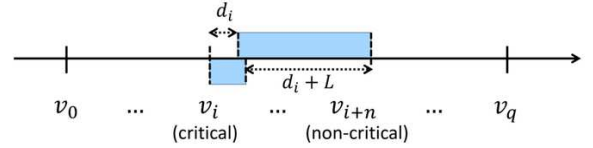}
    \caption{Time window adjustment for arrival vertex.}
    \label{fig:TW_Arrival}
\end{figure}

The DARP formulation given by \cite{2} considers either in-bound or out-bound request. Here, one vertex is always critical, which has narrower time window and the other is non-critical with time window usually set to be between $[0,T]$, where $T$ denotes the end of the day. In DARP, the ride time constraint is important for high user satisfaction. The service for a request starts at time $B_i$. The service time at the pick-up and drop-off point is indicated by $d_i$. The ride time of the user $L_i$ as shown in Fig.~\ref{fig:TW_Ridetime} is bounded by the ride time constraint $L$. 
In this work, a new time window adjustment (TW) method is proposed to improve the convergence of the tabu search heuristic towards global optimal solution. The methodology is illustrated using Figs.~\ref{fig:TW_Departure} and \ref{fig:TW_Arrival}. Initially, the service of this request (out-bound) consist of relaxed time window (at pick up point). When the time window $[0,T]$ of non critical vertex is constrained as per the proposed methodology, the direction of search is intensified towards more feasible region in the search space.

To adjust the time window of departure vertex (non critical), the earliest service time of departure vertex $e_i$ should not be earlier than $d_i+L$ from earliest service time of arrival vertex $e_{i+n}$; the latest service time of departure vertex $l_i$ should not be later than $l_{i+n}-d_i$. The adjustment made to departure vertex are given as follows:
\begin{align}
e_{i,new} &= \max\{e_{i,old},e_{i+n}-d_i-L\},
\label{eq:TWA_departure_1} \\
l_{i,new} &= \min\{l_{i,old},l_{i+n}-d_i\}.
\label{eq:TWA_departure_2}
\end{align}

In Eq.~\ref{eq:TWA_departure_1} and~\ref{eq:TWA_departure_2}, $e_{i,old}$ (=$0$) and $l_{i,old}$ (=$T$) are the earliest and latest time windows associated with the departure vertex that are adjusted to obtain $e_{i,new}$ and $l_{i,new}$ respectively.

Similarly, to adjust the time window of arrival vertex, the earliest service time of departure vertex $e_{i+n}$ should not be earlier than $e_i+d_i$, while the latest service time of departure vertex $l_{i+n}$ should not be later than $l_i+L+d_i$. The adjustments made to arrival vertex are given by Eq.~\ref{eq:TWA_arrival_1} and~\ref{eq:TWA_arrival_2}.


\begin{align}
e_{i+n,new} &= \max\{e_{i+n,old},e_i+d_i\},
\label{eq:TWA_arrival_1} \\
l_{i+n,new} &= \min\{l_{i+n,old},l_i+d_i+L\}.
\label{eq:TWA_arrival_2}
\end{align}

In Eq.~\ref{eq:TWA_arrival_1} and~\ref{eq:TWA_arrival_2}, $e_{i+n,old}$ (=$0$) and $l_{i+n,old}$ (=$T$) are the earliest and latest time windows associated with the arrival vertex that are adjusted to obtain $e_{i+n,new}$ and $l_{i+n,new}$ respectively.

The improved tabu search (ITS) is obtained by incorporating the two proposed techniques, i.e., construction heuristic (CH) and time window adjustment (TW) into the tabu search variant $TS_{32}$. ITS ($TS_{32(CH+TW)}$) is tested against the benchmark of \cite{6} and the simulation results are presented in the next section.


\section{Simulation Results}
\label{sec:simulation}

\begin{figure*}[htp]
    \begin{centering}
		\subfigure[R1a test instance]{\includegraphics[scale=0.156]{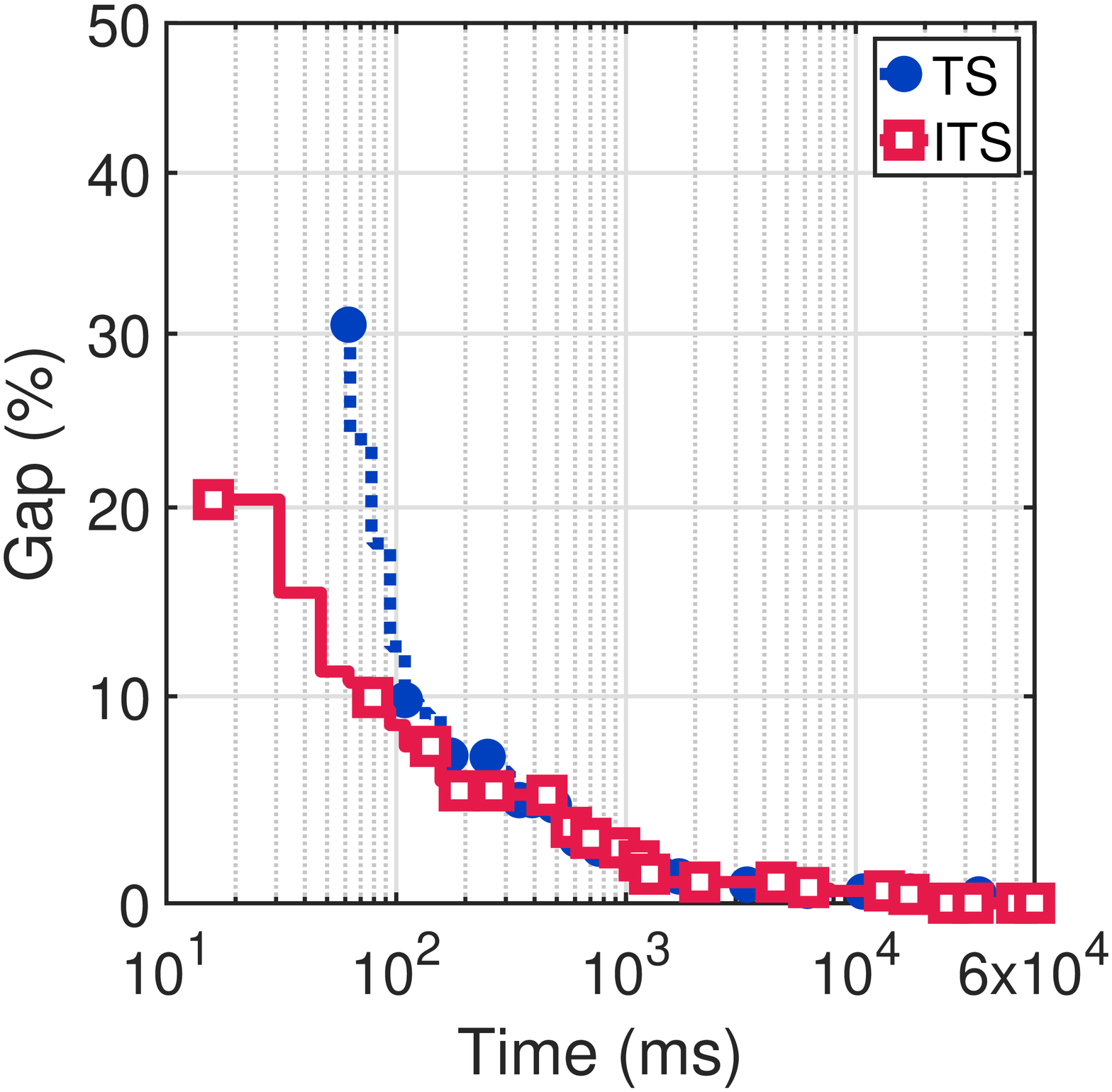}\label{fig:R1a_1}}
		\subfigure[R1b test instance]{\includegraphics[scale=0.156]{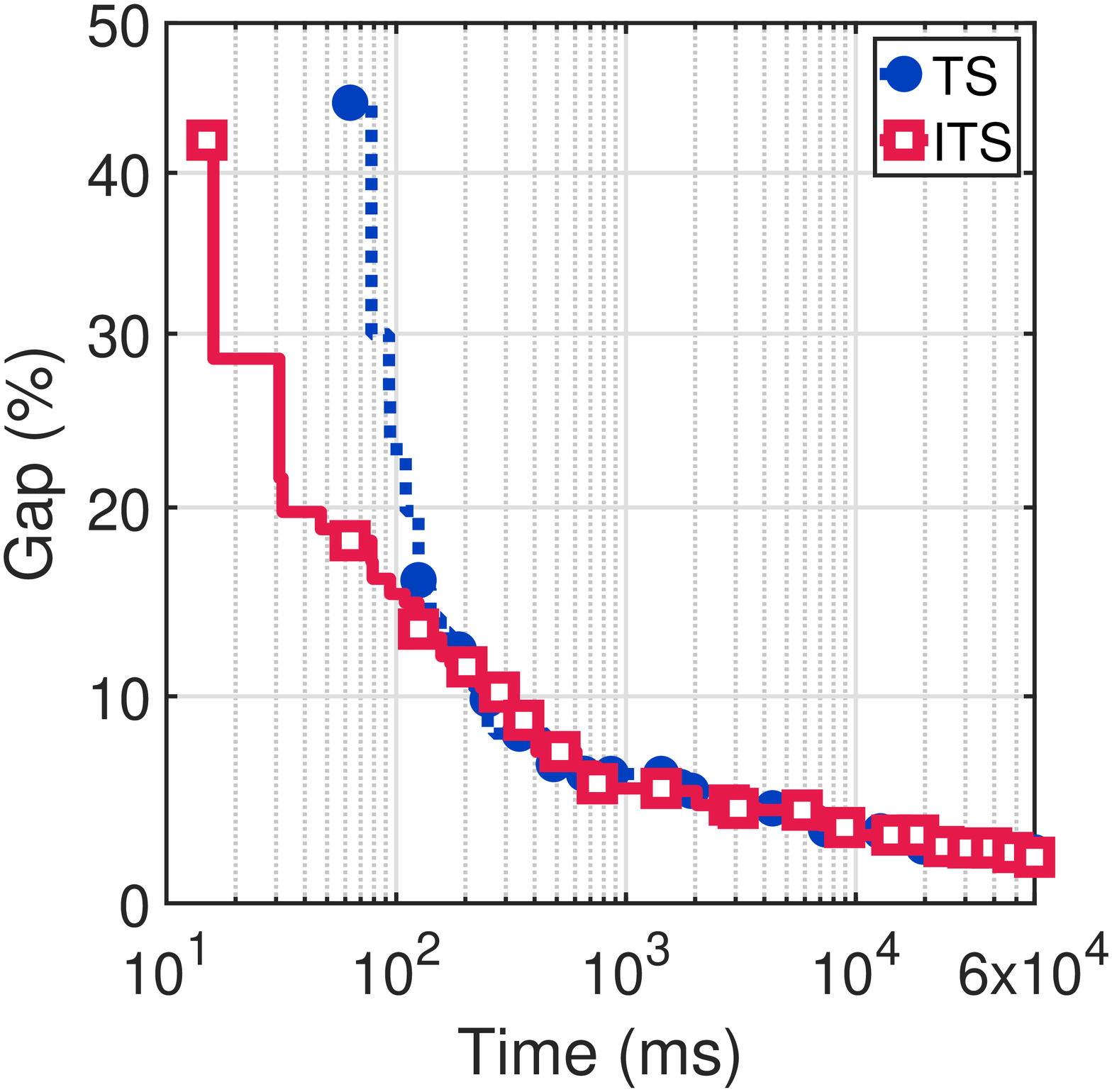}\label{fig:R2a_1}}
		\subfigure[R2a test instance]{\includegraphics[scale=0.156]{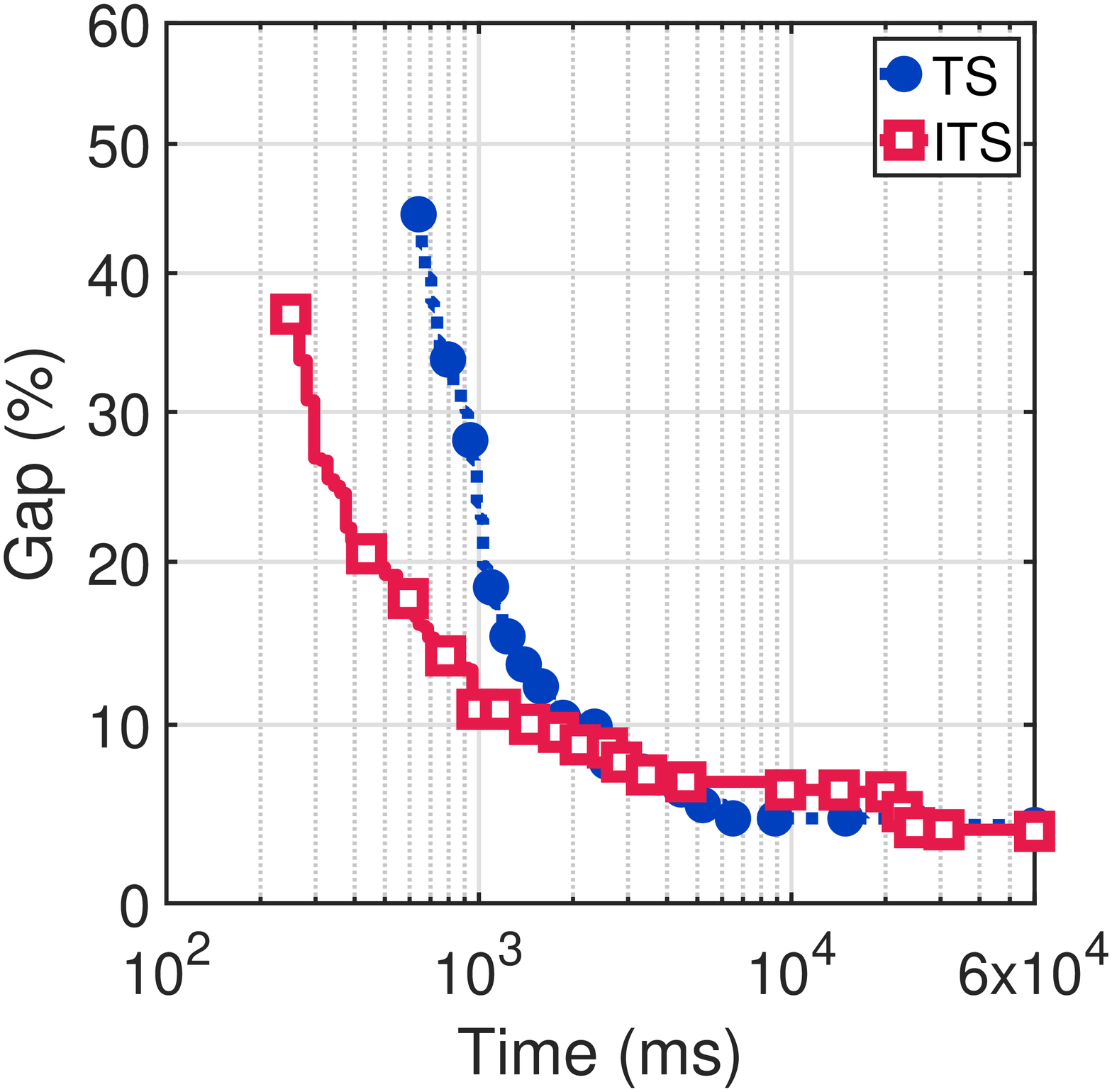}\label{fig:R3a_1}}
		\subfigure[R2b test instance]{\includegraphics[scale=0.156]{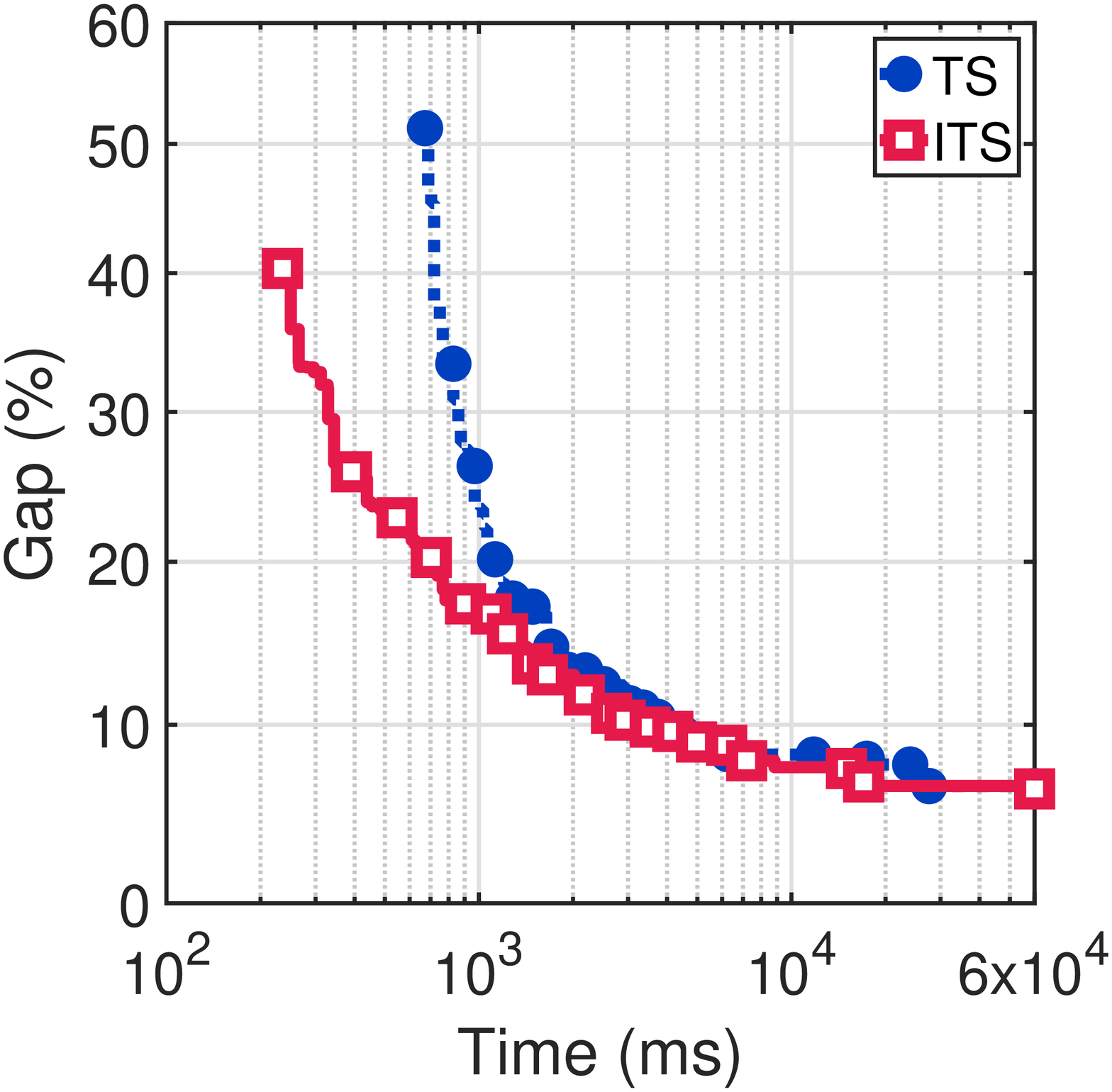}\label{fig:R4a_1}}
		\subfigure[R3a test instance]{\includegraphics[scale=0.156]{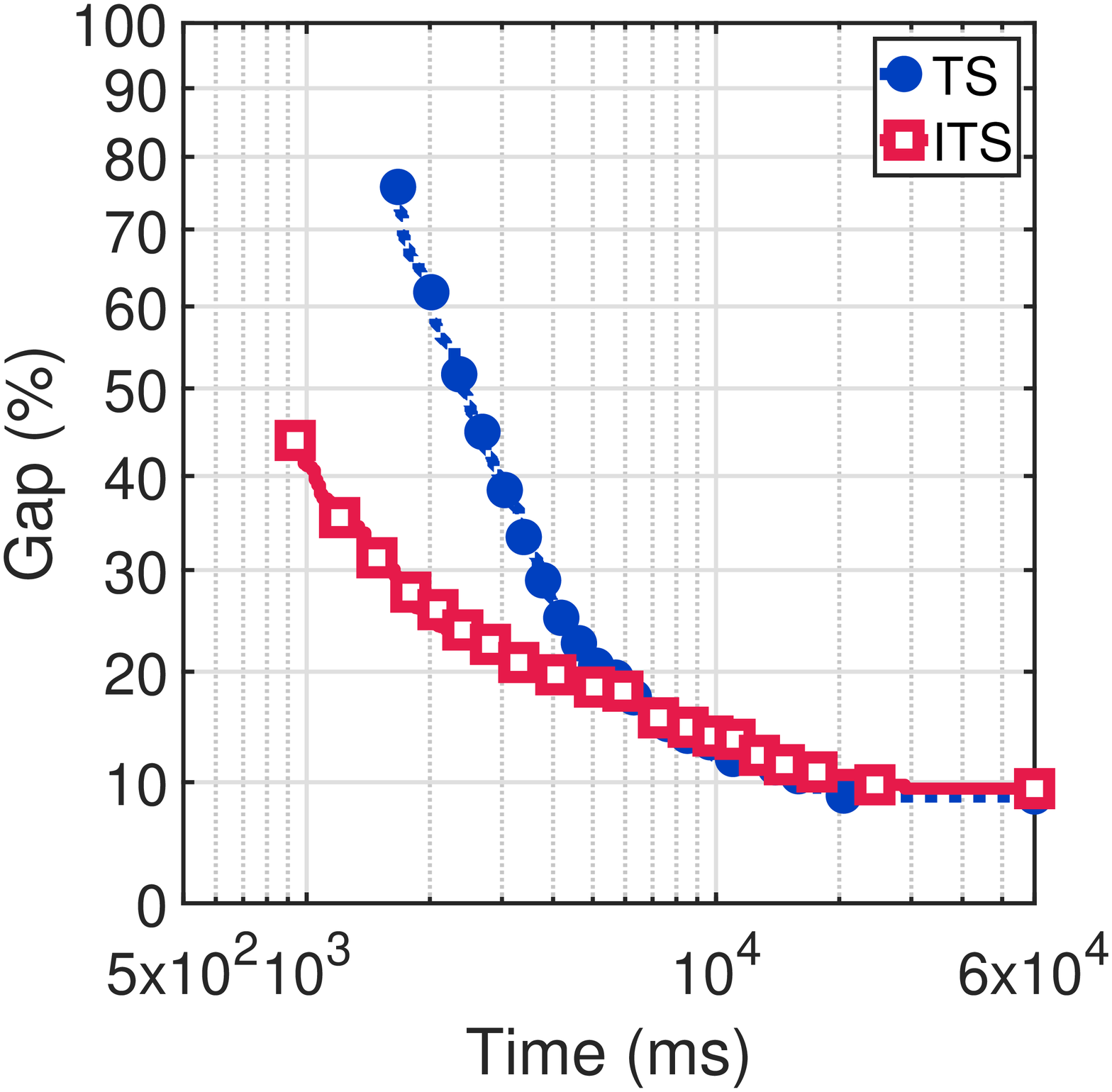}\label{fig:R5a_1}}
		\subfigure[R3b test instance]{\includegraphics[scale=0.156]{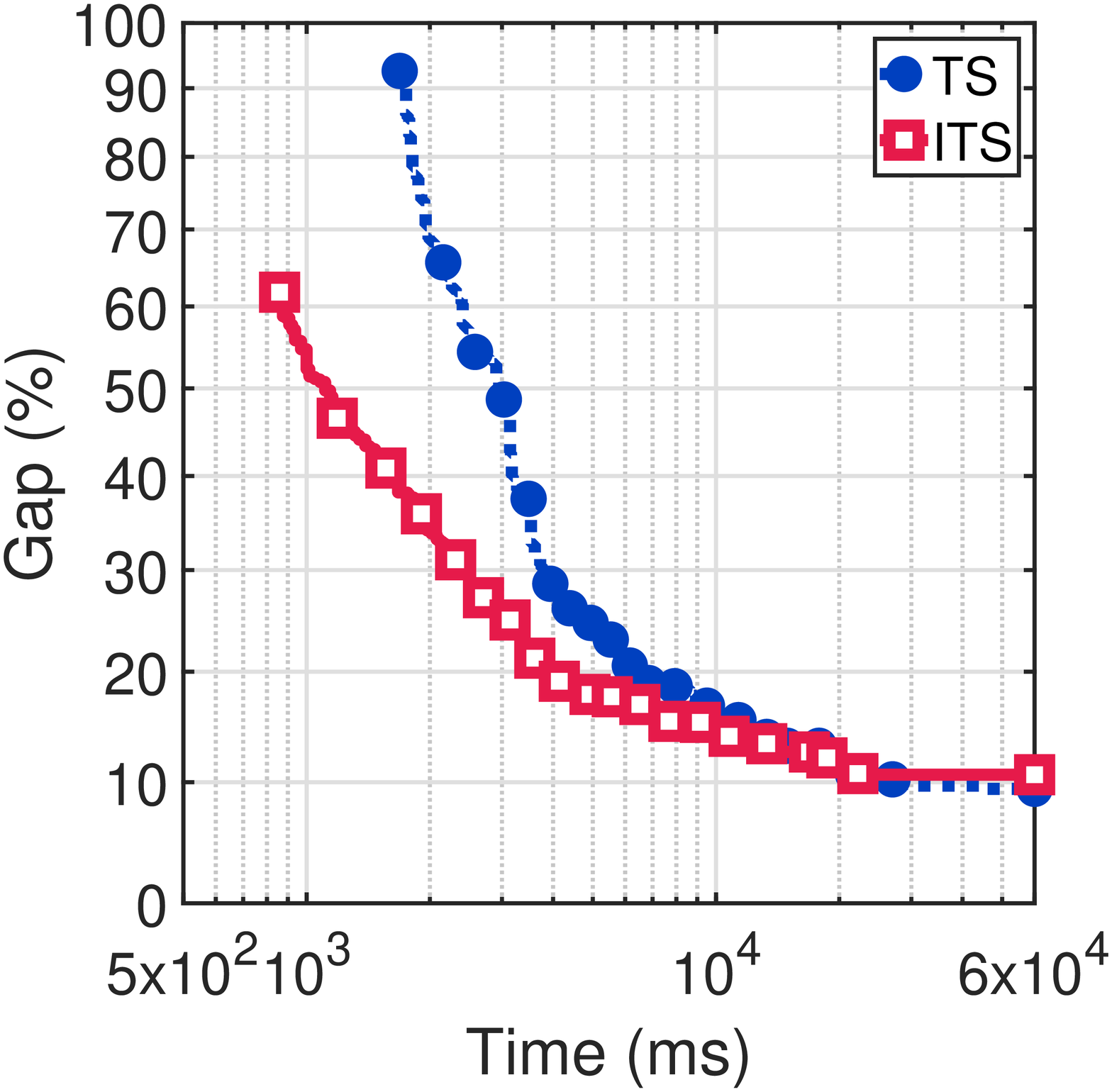}\label{fig:R6a_1}}
		\subfigure[R4a test instance]{\includegraphics[scale=0.156]{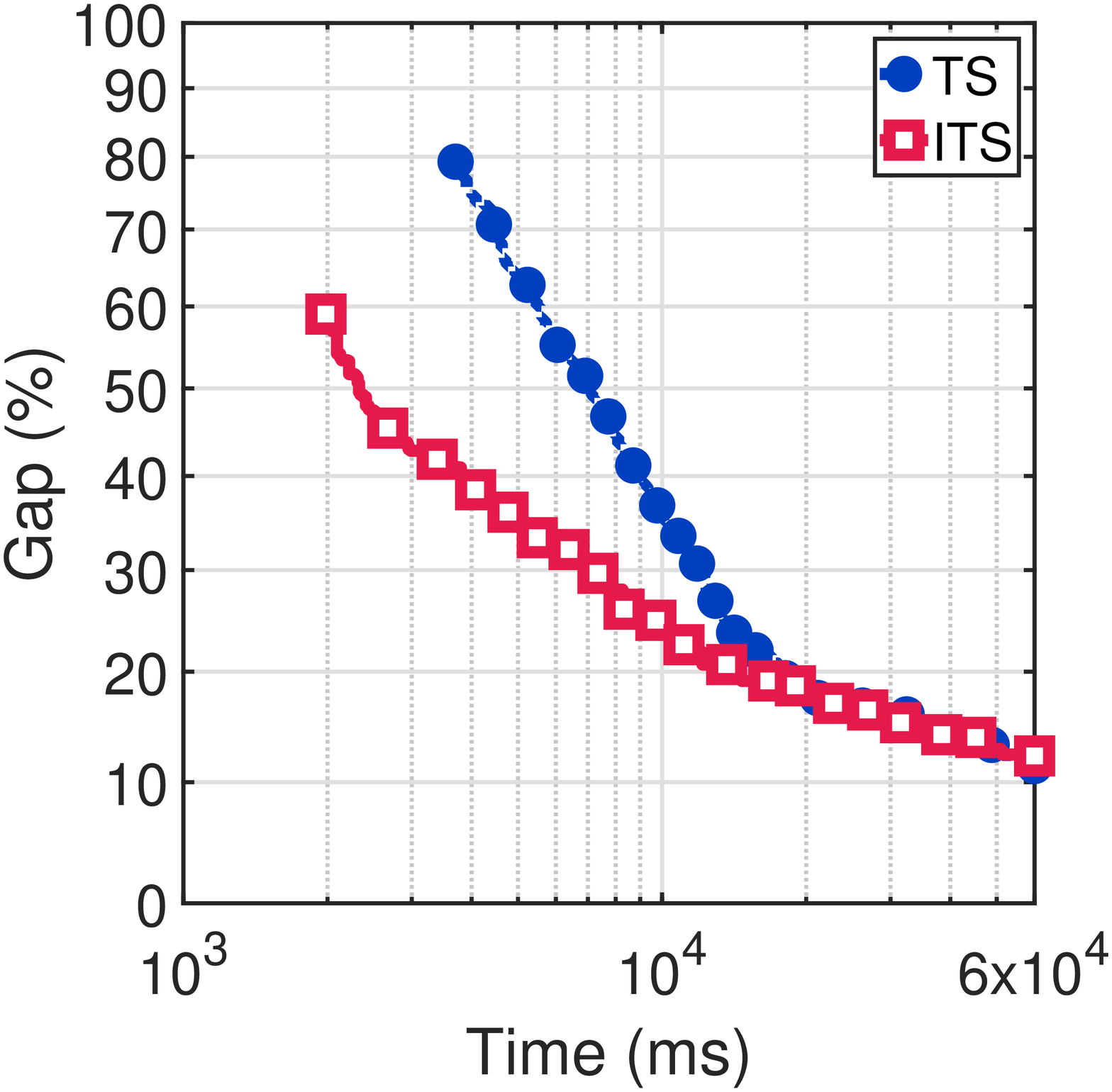}\label{fig:R7a_1}}
		\subfigure[R4b test instance]{\includegraphics[scale=0.156]{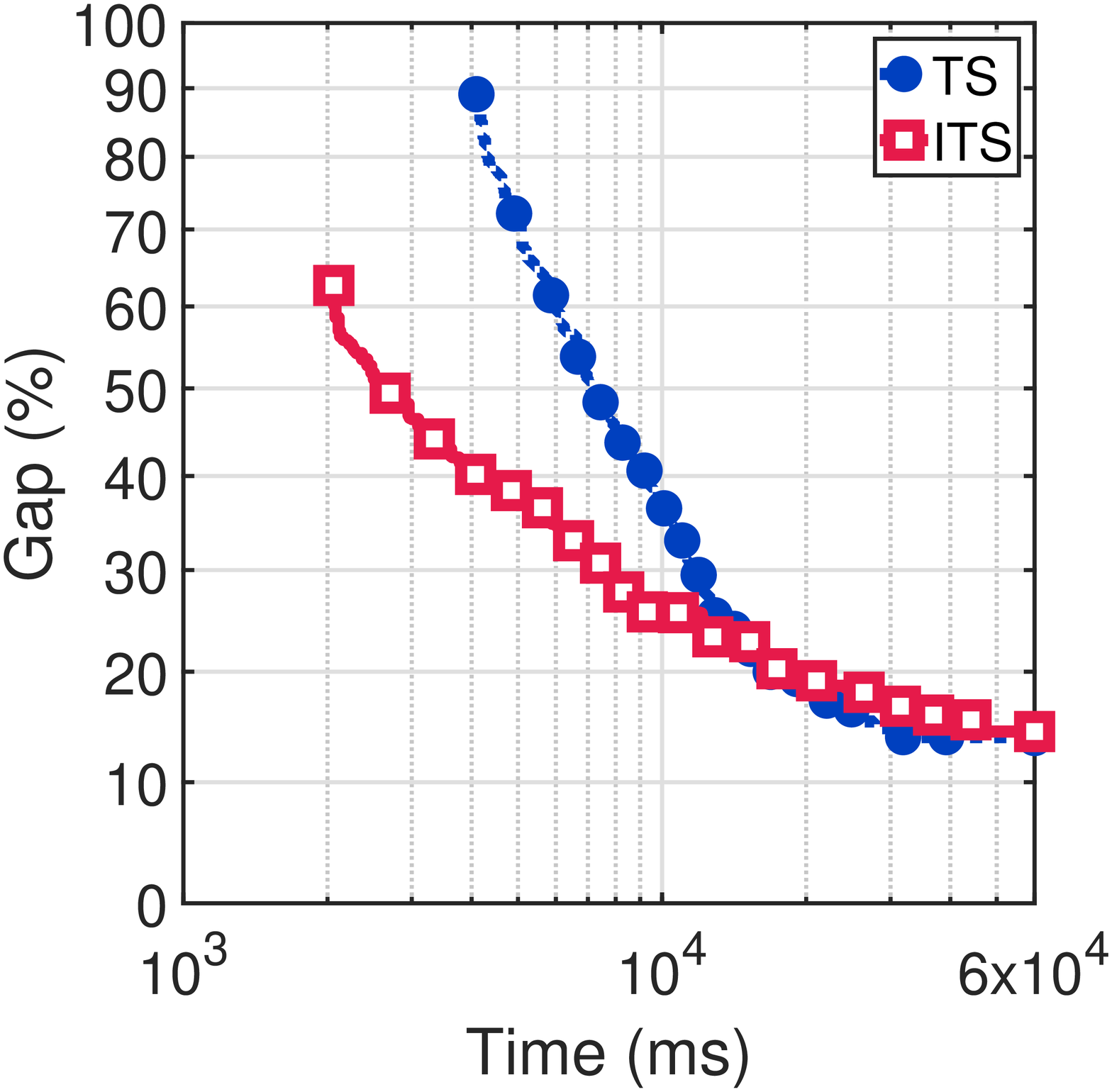}\label{fig:R8a_1}}
		\subfigure[R5a test instance]{\includegraphics[scale=0.156]{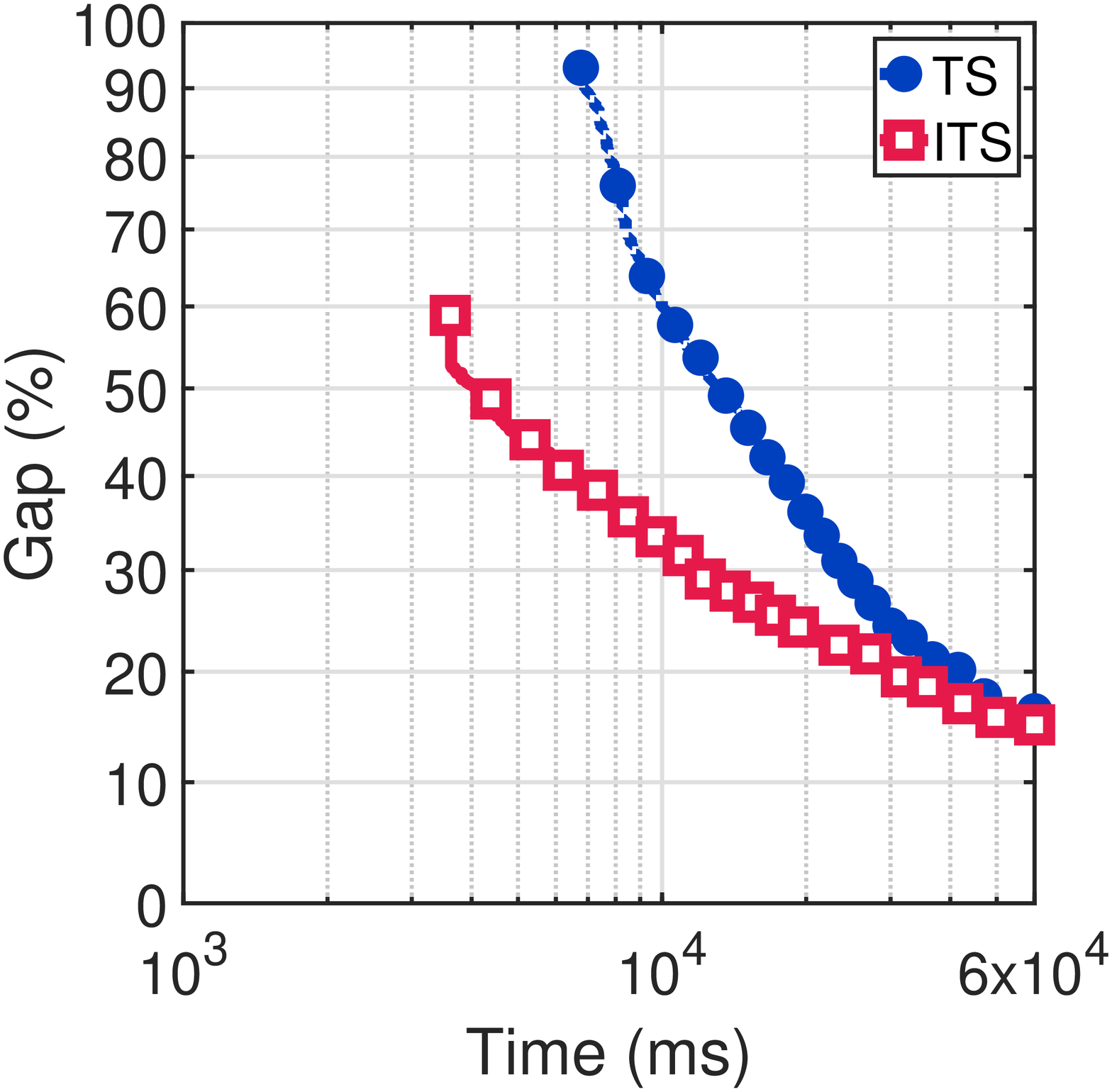}\label{fig:R9a_1}}
		\subfigure[R5b test instance]{\includegraphics[scale=0.156]{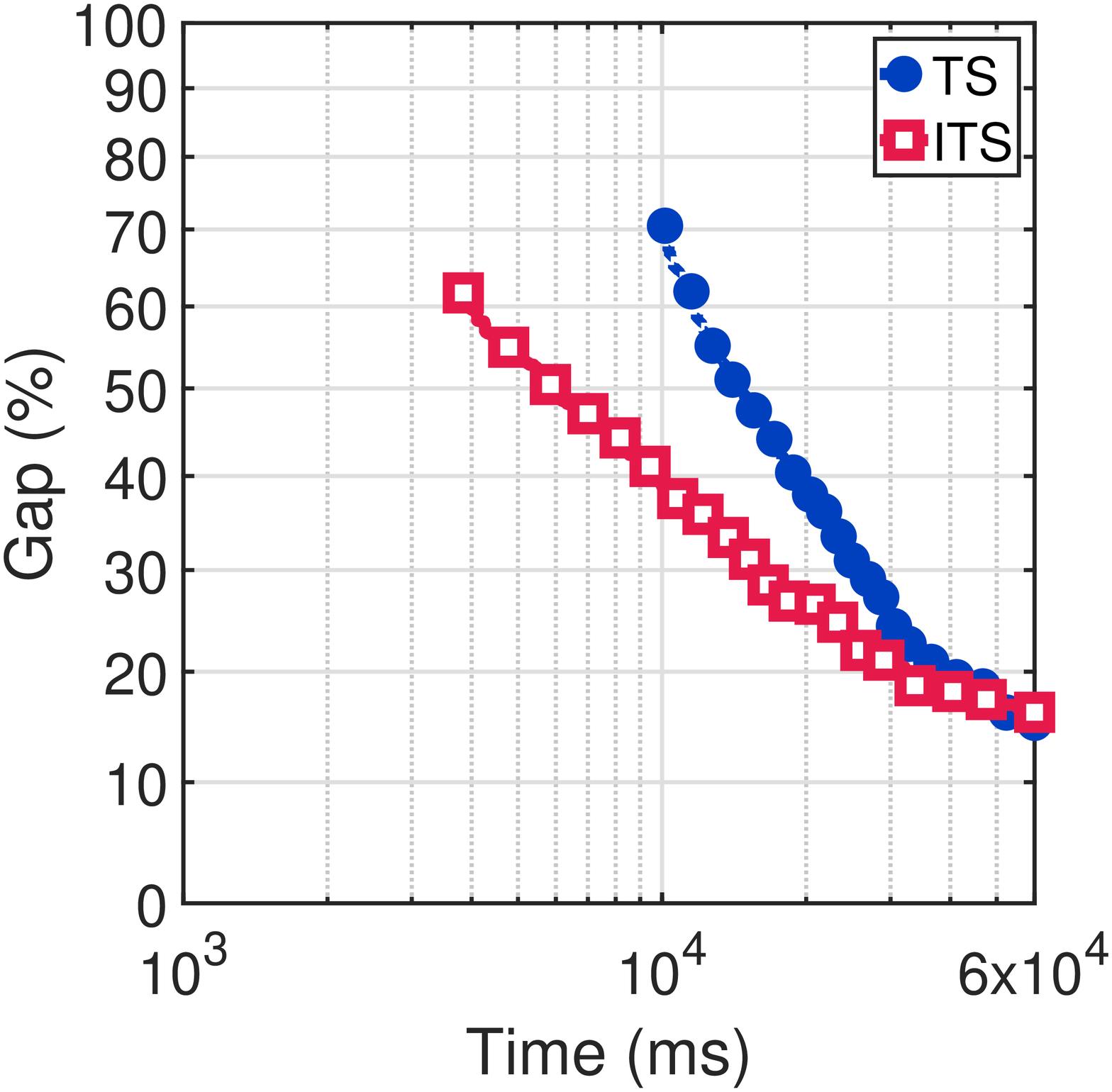}\label{fig:R10a_1}}
		\subfigure[R6a test instance]{\includegraphics[scale=0.156]{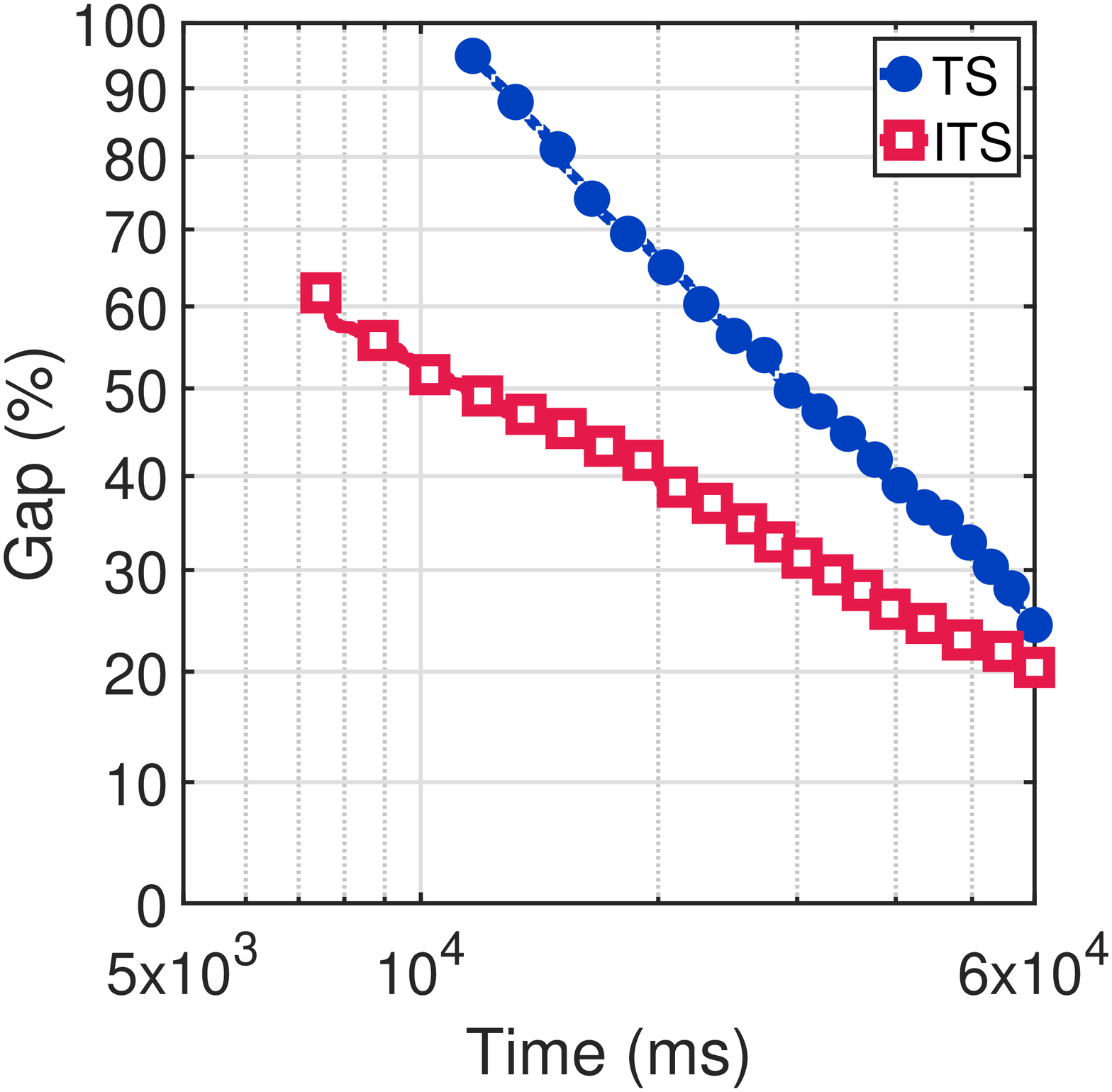}\label{fig:R1b_1}}
		\subfigure[R6b test instance]{\includegraphics[scale=0.156]{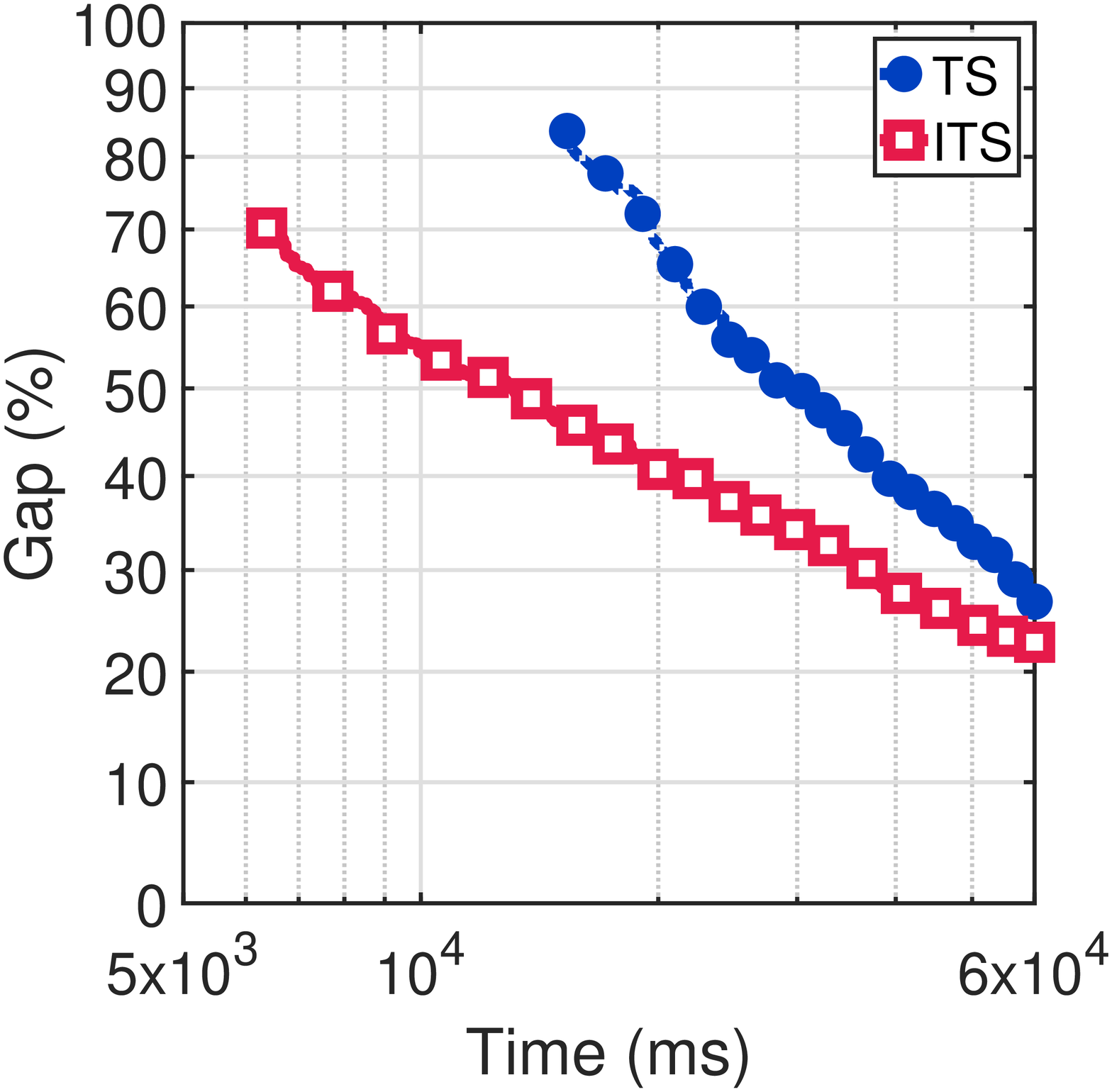}\label{fig:R2b_1}}
        \subfigure[R7a test instance]{\includegraphics[scale=0.156]{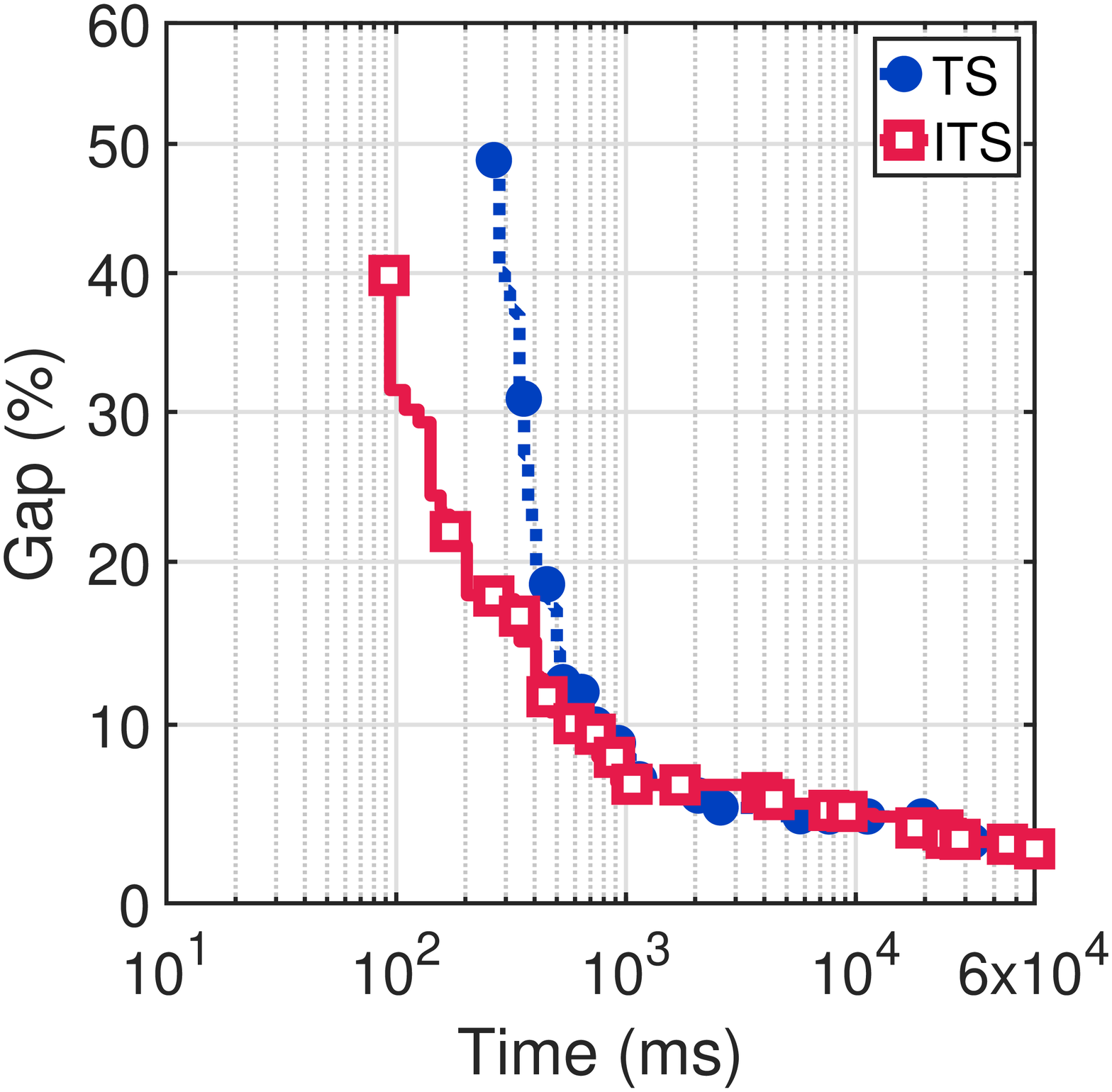}\label{fig:R3b_1}}
		\subfigure[R7b test instance]{\includegraphics[scale=0.156]{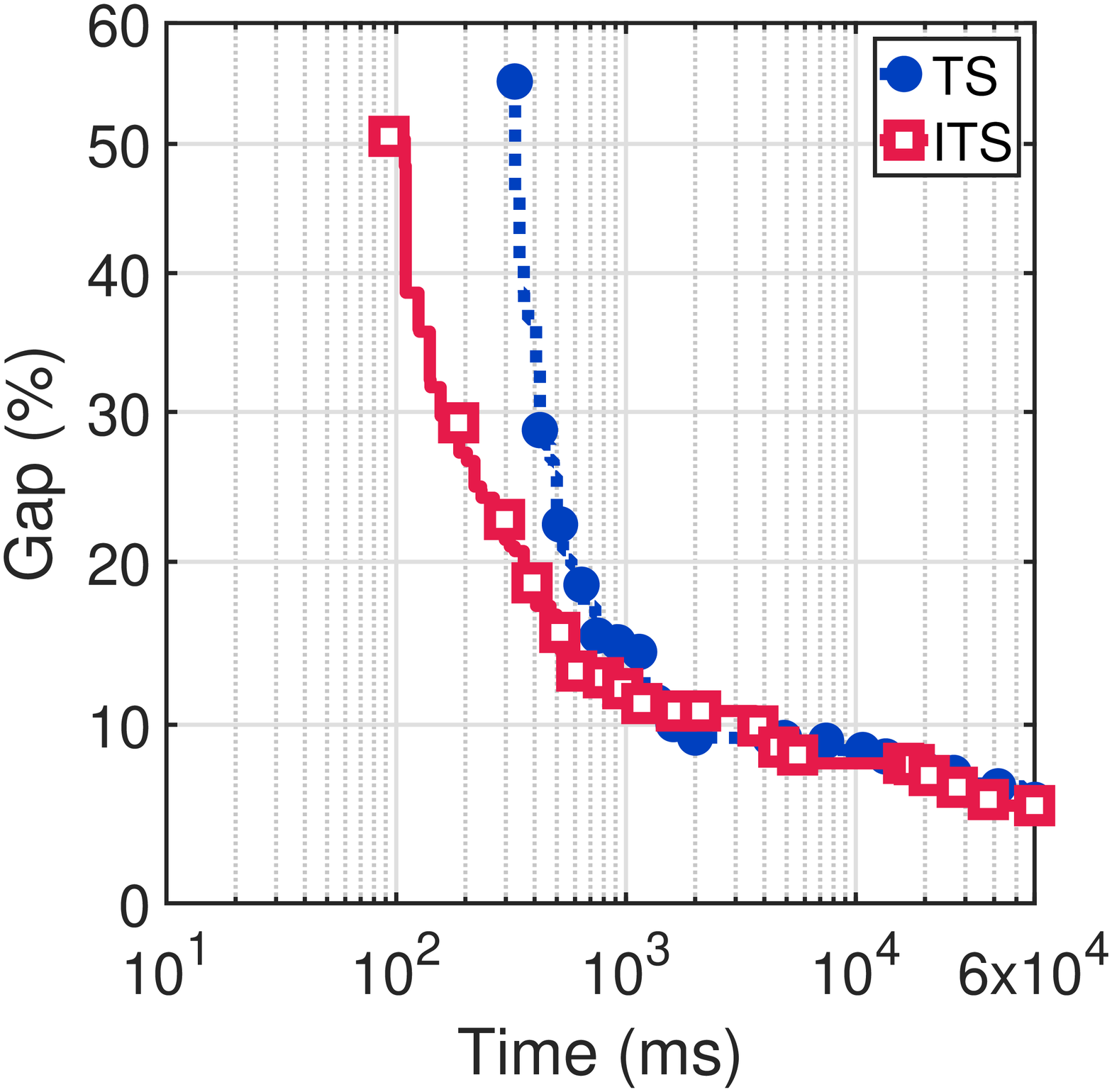}\label{fig:R4b_1}}
		\subfigure[R8a test instance]{\includegraphics[scale=0.156]{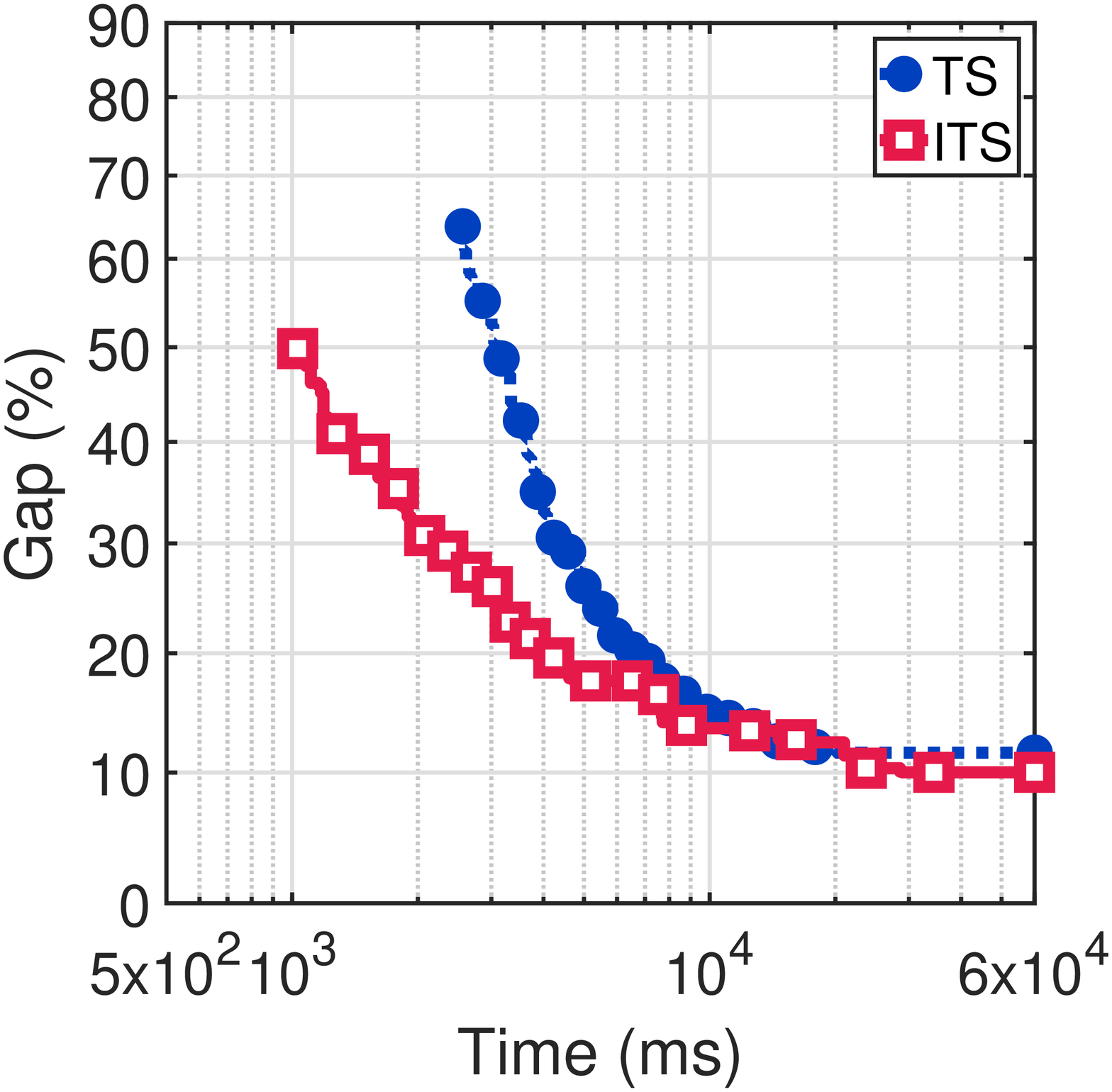}\label{fig:R5b_1}}
		\subfigure[R8b test instance]{\includegraphics[scale=0.156]{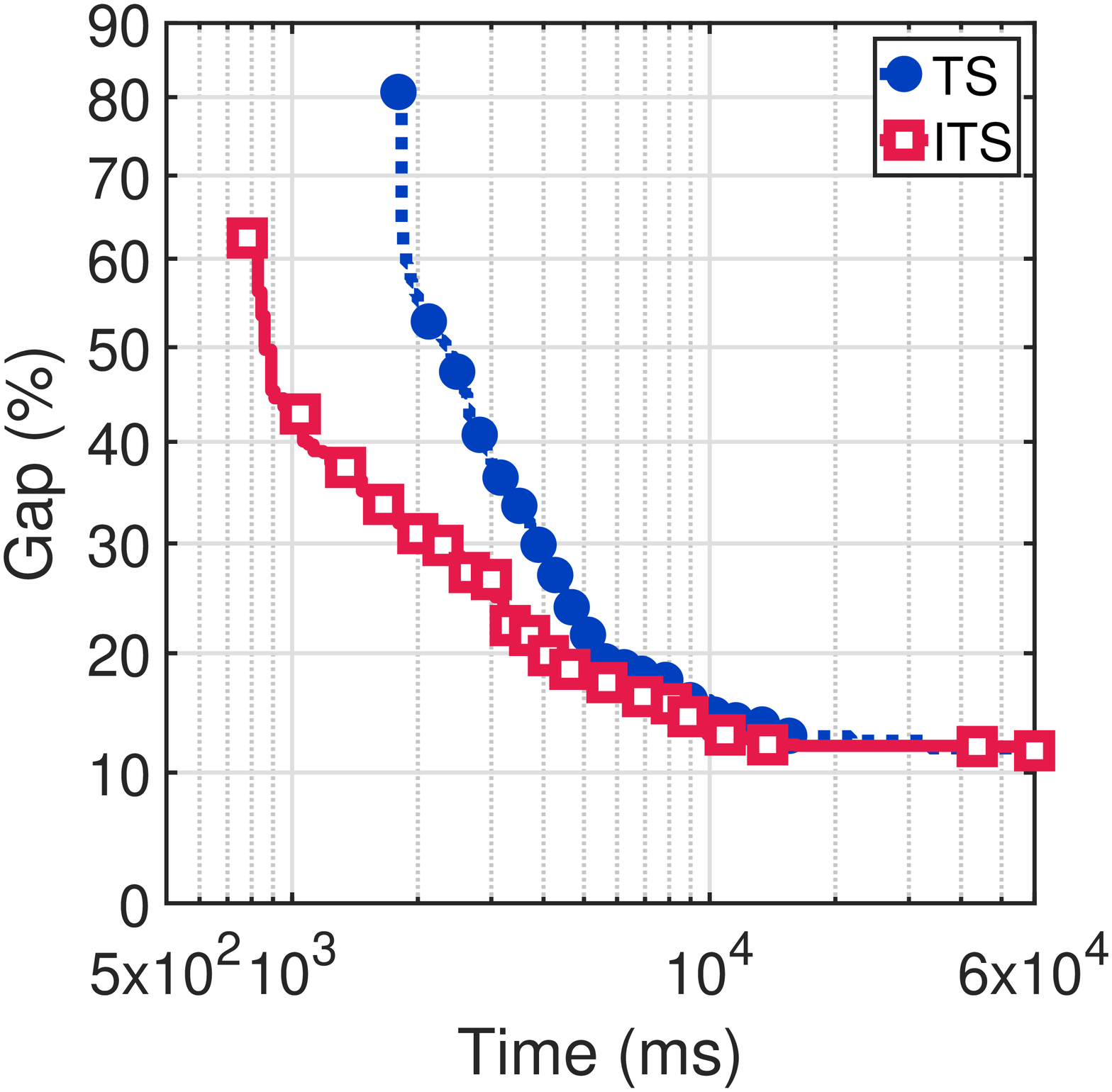}\label{fig:R6b_1}}
		\subfigure[R9a test instance]{\includegraphics[scale=0.156]{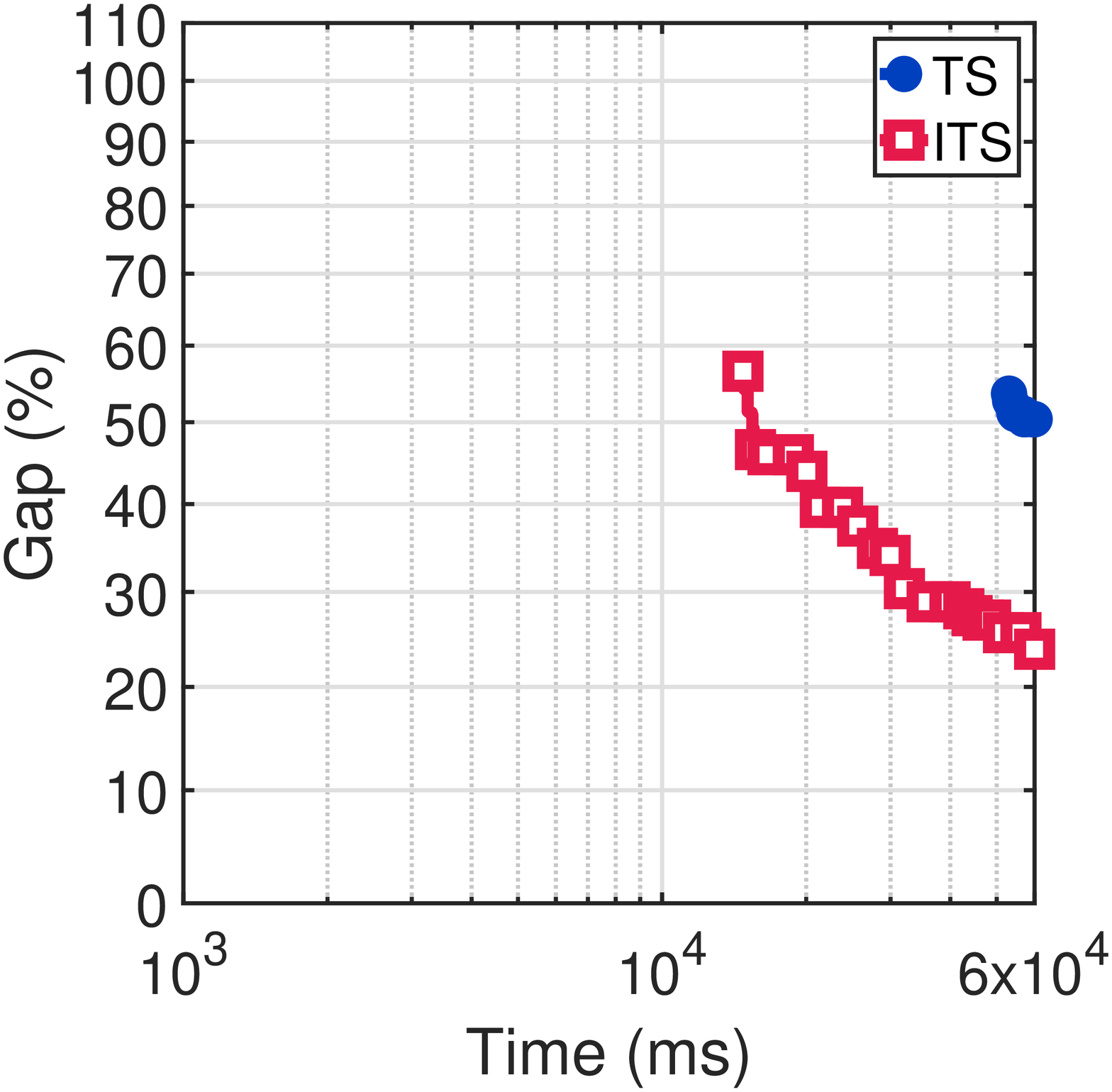}\label{fig:R7b_1}}
		\subfigure[R9b test instance]{\includegraphics[scale=0.156]{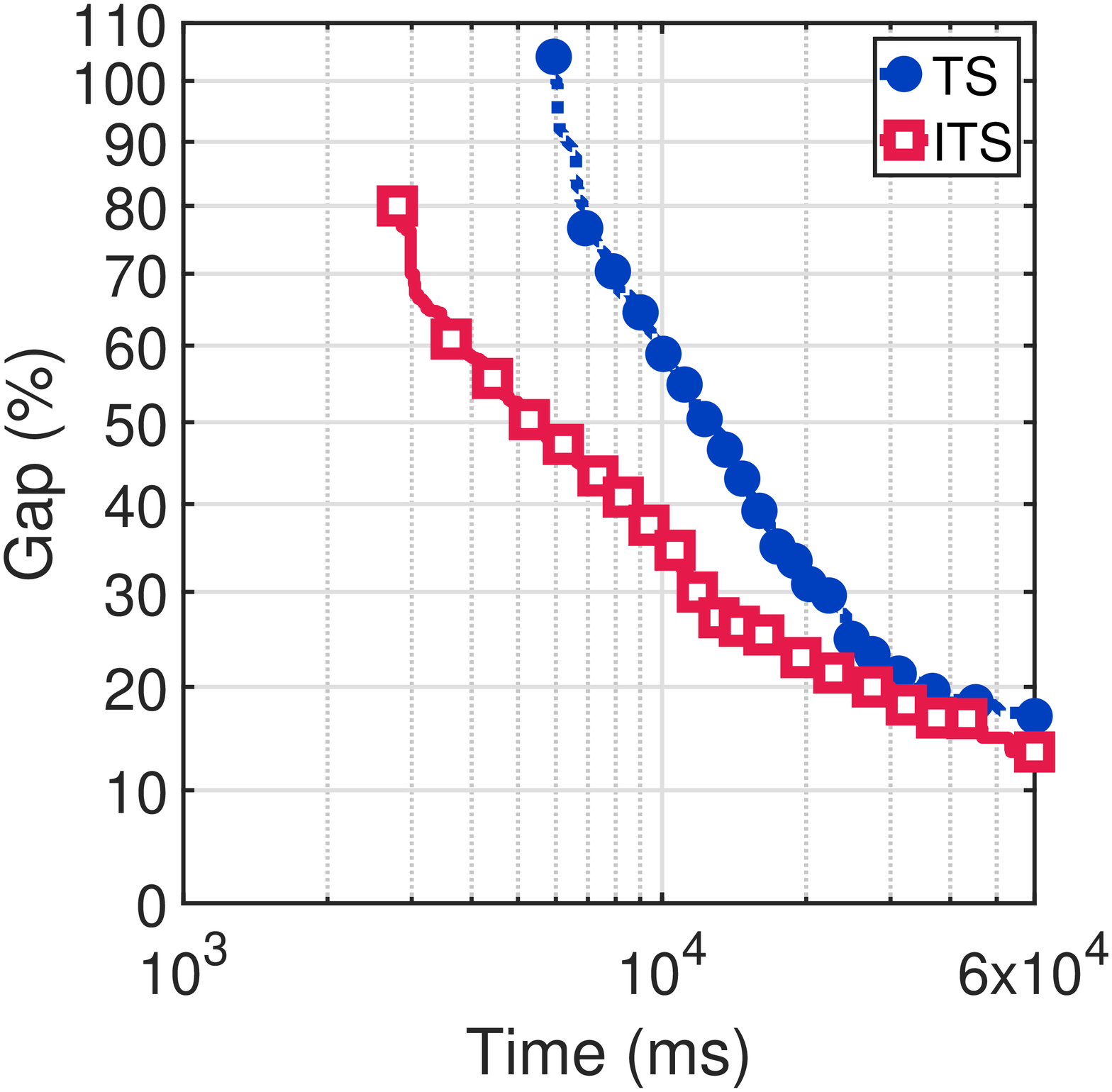}\label{fig:R8b_1}}
		\subfigure[R10a test instance]{\includegraphics[scale=0.156]{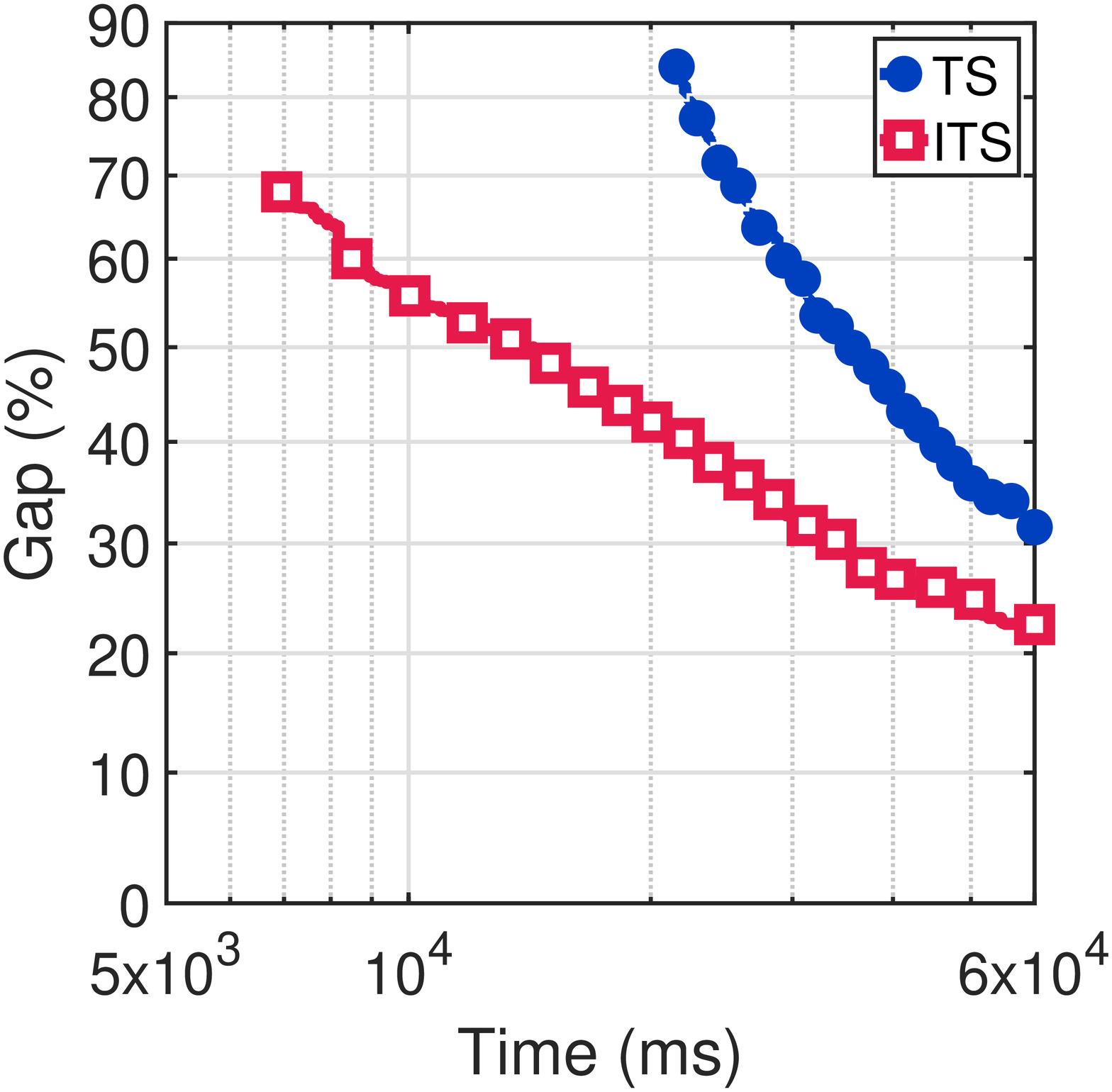}\label{fig:R9b_1}}
		\subfigure[R10b test instance]{\includegraphics[scale=0.156]{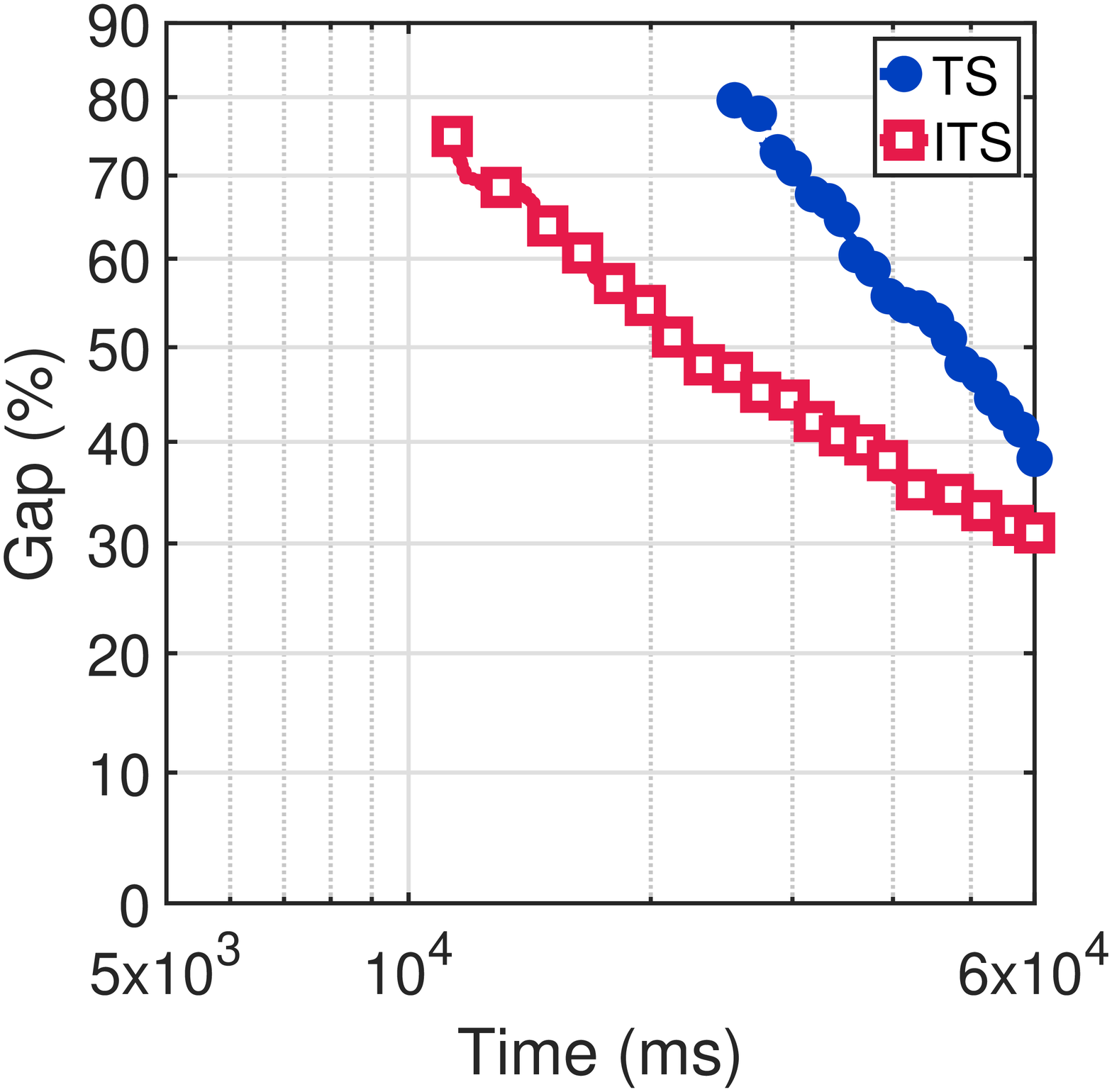}\label{fig:R10b_1}}
		\caption{Convergence analysis for improved tabu search (ITS) w.r.t. tabu search (TS) \cite{2} using various test instances.}
		\label{fig:Convergence_Analysis}
	\end{centering}
\end{figure*}

\begin{table*}[htp]
\begin{subtable}{}
\centering
{\scriptsize
\begin{tabular}{>{}c*{14}{c}}\toprule
    \multirow{2}{*}{Test Instance} & \multirow{2}{*}{BKS} & \multicolumn{6}{c}{Tabu Search (TS) (\cite{2})} & \multicolumn{6}{c}{Improved Tabu Search (ITS)} \\\cmidrule(lr){3-8}\cmidrule(lr){9-14}
                   &          & 1 sec & 2 sec & 5 sec & 15 sec & 30 sec & 60 sec  & 1 sec & 2 sec & 5 sec & 15 sec & 30 sec & 60 sec \\ \midrule
    R1a	& 190.02 \cite{17} &	193.26	&191.66	&191.11	&191.05	&190.79	&190.02	&193.77	&191.88&	191.66&	191.11	&190.02	&190.02 \\
    R2a	& 301.34 \cite{17}	&373.23	&332.21	&318.96	&315.31	&314.20	&314.18	&334.22	&328.48	&321.53	&320.11	&313.40	&313.13 \\
    R3a	&532.00 \cite{17}	&-	&863.26	&644.51	&592.15	&578.66&	578.66	&750.64&	670.69&	632.84&	593.06&	582.25&	582.25 \\
    R4a	&570.25 \cite{17}	&-	&-	&935.45	&698.30	&661.68	&635.35	&-	&906.90&	770.07&	679.35	&658.53	&640.39 \\
    R5a	&626.93	\cite{6}&-	&-	&-	&918.30	&780.16	&729.24&	-&	-	&906.06	&795.77&	753.48&	721.48 \\
    R6a	&785.26 \cite{17}	&-	&-	&-	&1416.25	&1171.76	&977.61	&-	&-	&-	&1143.63	&1034.10	&945.55 \\
    R7a	&291.71 \cite{17}	&315.87	&310.79	&305.58	&305.55	&302.81	&300.68	&310.89	&310.74	&307.63	&305.52	&301.83	&300.30 \\
    R8a	&487.84 \cite{17}	&-	&-	&614.78	&548.96	&544.45	&544.45	&-	&642.78	&574.11	&549.66	&536.74	&536.74 \\
    R9a	&658.31 \cite{17} &	-&	-&	-&	-&	-&	989.98&	-&	-&	-&	1013.78&	882.53&	815.50 \\
    R10a &851.82 \cite{6} &	-&	-&	-&	-&	1357.03&	1120.59&	-&	-&	-&	1262.90	&1128.65&	1043.75 \\
    R1b	&164.46 \cite{17}	&174.55&	173.21&	171.81&	169.14&	168.80&	168.35&	173.39&	173.19&	171.67	&169.71	&168.70	&167.98 \\
    R2b	&295.66 \cite{17}	&366.00	&334.91	&322.87	&320.08	&314.75	&314.75	&345.04	&334.01&	322.25&	317.70	&314.76	&314.28 \\
    R3b	&484.83 \cite{17}	&-	&816.63	&604.61	&548.84	&531.89	&530.54	&738.39	&650.30	&571.50	&546.19	&536.46	&536.46 \\
    R4b	&529.33 \cite{17}	&-	&-	&904.96	&651.59	&607.58&	602.90&	-&	-&	731.19&	651.49&	625.01	&605.95 \\
    R5b	&577.29	\cite{6} &-	&-	&-	&859.20	&724.33	&665.43	&-	&-	&888.87	&759.56	&698.75	&670.99 \\
    R6b	&730.69	\cite{6} &-	&-	&-	&-	&1096.04	&926.58	&-	&-	&-	&1068.33	&980.66	&897.41 \\
    R7b	&248.21 \cite{17}	&285.32	&271.17	&270.71	&267.24	&264.60	&262.29	&276.14	&275.04	&268.66	&267.47	&262.76	&261.49 \\
    R8b	&458.73	\cite{6} &-	&711.44	&559.40	&518.31	&516.46	&513.56	&655.42	&600.62	&541.42	&514.95	&514.55	&512.68 \\
    R9b	&593.49 \cite{17}	&-	&-	&-	&842.19	&720.25	&694.87	&-	&-	&903.30	&748.44	&712.13	&674.22 \\
    R10b &785.68 \cite{6} &-	&-	&-	&-	&1342.78	&1086.40	&-	&-	&-	&1287.53	&1132.89	&1029.22 \\
\bottomrule
\end{tabular}}
\caption{Comparison of the median of travel cost: Tabu Search (TS) (\cite{2}) vs Improved Tabu Search (ITS).}
\label{tab:comparison3}
\end{subtable}
\bigskip
\bigskip
\begin{subtable}{}
\centering
{
\begin{tabular}{>{}c*{8}{c}}\toprule
    \multirow{3}{*}{Test Instance} & \multirow{3}{*}{BKS} & \multicolumn{6}{c}{First Feasible Solution} \\\cmidrule(lr){3-8}
        &    & \multicolumn{3}{c}{Tabu Search (TS) (\cite{2})} & \multicolumn{3}{c}{Improved Tabu Search (ITS)} \\\cmidrule(lr){3-5}\cmidrule(lr){6-8}
        &        & Cost & Gap (\%) & Time (ms) & Cost & Gap (\%) & Time (ms) \\ \midrule
    R1a	& 190.02 \cite{17} & 248.05&	30.54&	62&	228.85	&20.43&	16 \\
    R2a	& 301.34 \cite{17} &  435.34	&44.47&	641&	412.74&	36.97&	250\\
    R3a	& 532.00 \cite{17} &  935.17&	75.78&	1672&	765.95&	43.98&	938\\
    R4a	& 570.25 \cite{17} &  1022.51&	79.31&	3704&	906.90&	59.04&	1985\\
    R5a	& 626.93 \cite{6} &  1210.45&	93.08&	6766&	996.00	&58.87&	3610\\
    R6a	& 785.26 \cite{17} &  1530.40&	94.89&	11642&	1269.81&	61.71&	7469\\
    R7a	& 291.71 \cite{17} &  433.79&	48.71&	266&	407.89&	39.83&	93\\
    R8a	& 487.84 \cite{17} &  799.18&	63.82&	2562&	731.00	&49.84&	1031\\
    R9a	& 658.31 \cite{17} &  1011.51&	53.65&	53051&	1030.78&	56.58&	1751\\
    R10a & 851.82 \cite{6}&  1567.82&	84.06&	21533&	1431.05&	68.00&	6954\\
    R1b	& 164.46 \cite{17} &  237.80&	44.59&	63&	233.73&	42.12&	15\\
    R2b	& 295.66 \cite{17} &  447.23&	51.26&	672&	414.94&	40.35&	235\\
    R3b	& 484.83 \cite{17} &  933.77&	92.60&	1688&	784.44	&61.80&	859\\
    R4b	& 529.33 \cite{17} &  1000.97&	89.10&	4094&	860.97&	62.65&	2062\\
    R5b	& 577.29 \cite{6} &  984.23&	70.49&	10141&	933.52&	61.71&	3844\\
    R6b	& 730.69 \cite{6} &  1342.25&	83.70&	15329&	1243.52&	70.18&	6376\\
    R7b	& 248.21 \cite{17} &  384.93&	55.08&	328&	373.82&	50.60&	93\\
    R8b	& 458.73 \cite{6} &  828.88&	80.69&	1797&	745.12&	62.43&	781\\
    R9b	& 593.49 \cite{17} &  1211.39&	104.11&	5938&	1068.24&	79.99&	2797\\
    R10b & 785.68 \cite{6} &  1411.05&	79.60&	25424&	1374.35&	74.92&	11329\\
\bottomrule
\end{tabular}}
\caption{Comparison of the initial travel cost: TS (\cite{2}) vs ITS.}
\label{tab:comparison2}
\end{subtable}
\end{table*}

The proposed ITS heuristic ($\text{TS}_{\text{32(CH+TW)}}$) is implemented in C++. Simulations have been carried out on a computer running 2.1 GHz Intel Xeon E5-2620 v4 processor with 128 GB RAM. The parameters suggested by \cite{2} have been followed to conduct a fair comparison of the results obtained using the proposed heuristic with the existing tabu search \cite{2} method. Alongside the variant of $\text{TS}_{\text{32}}$; $\text{TS}_{\text{32(CH+TW)}}$, a hybrid of $\text{TS}_{\text{32}}$ with both construction heuristic and time window adjustment have been tested. 


In Figs. \ref{fig:R1a_1} - \ref{fig:R10b_1}, we depict the progression of the solution towards the benchmark during the first sixty seconds of execution. In these plots, the blue and red lines correspond to the Cordeau's tabu search \cite{2} and the proposed ITS heuristic respectively. The plots show the median of gap obtained from fifteen independent simulations run for sixty seconds. From these plots, it is clearly evident that the proposed improved tabu search (ITS) heuristic outperforms tabu search (TS) \cite{2} for all the DARP test instances.

The following inference is made from these numerical experiments: Construction heuristic (CH) along with the time window adjustment (TW) not only contributes to significant speed up in the convergence, but also finds good feasible solution in a shorter time.

In Table \ref{tab:comparison3}, we compare the travel cost of TS and ITS at various time instances \{1, 2, 5, 15, 30, 60\} sec for the benchmark instances. The presentation style is adopted from \cite{13}. It is observed that the solution quality of ITS is always better when compared to TS, especially during the first three minutes of the execution. Based on the best known solutions listed in the second column labeled with `BKS' in Table \ref{tab:comparison3}, it can be concluded that the proposed method attains near optimal solutions in shorter time. As mentioned in Section \ref{sec:intro}, the proposed construction heuristic of the improved tabu search (ITS) helps to produce a good feasible initial solution rapidly. Table \ref{tab:comparison2} provides an empirical evidence for this claim and presents a comparison of first feasible solutions for various test instances and the time at which they are obtained. The next section concludes the paper.

\section{Conclusion}
\label{sec:conclusion}
In this paper, an improved tabu search heuristic has been proposed to solve static dial-a-ride problem. Several variants of tabu search method based on the neighborhood evaluation and insertion techniques have been tested to analyse the convergence behavior. In the existing tabu search heuristic, two performance bottle necks have been identified: i) longer run time requirement to obtain first feasible solution and ii) slower convergence to the global optimum. To address these, two new techniques, i.e., construction heuristic and time window adjustment have been proposed to improve the performance. Simulation results for various test instances show that the proposed ITS heuristic not only improves the convergence, but also finds high quality solution faster. Moreover, the approach can be extended to dynamic DARP. Some other possible directions for future work could be the parallelization of the developed tabu search algorithm using GPUs to reduce the computation time. 

\bibliographystyle{IEEEtran}
\bibliography{Tabu_Search}

\begin{thebibliography}{10}
\providecommand{\url}[1]{#1}
\csname url@samestyle\endcsname
\providecommand{\newblock}{\relax}
\providecommand{\bibinfo}[2]{#2}
\providecommand{\BIBentrySTDinterwordspacing}{\spaceskip=0pt\relax}
\providecommand{\BIBentryALTinterwordstretchfactor}{4}
\providecommand{\BIBentryALTinterwordspacing}{\spaceskip=\fontdimen2\font plus
\BIBentryALTinterwordstretchfactor\fontdimen3\font minus
  \fontdimen4\font\relax}
\providecommand{\BIBforeignlanguage}[2]{{%
\expandafter\ifx\csname l@#1\endcsname\relax
\typeout{** WARNING: IEEEtran.bst: No hyphenation pattern has been}%
\typeout{** loaded for the language `#1'. Using the pattern for}%
\typeout{** the default language instead.}%
\else
\language=\csname l@#1\endcsname
\fi
#2}}
\providecommand{\BIBdecl}{\relax}
\BIBdecl

\bibitem{1}
L.~Corpuz-Bosshart, \emph{Road pricing most effective in reducing vehicle
  emissions}.\hskip 1em plus 0.5em minus 0.4em\relax UBC News, The University
  of British Columbia, October, 2017.

\bibitem{15}
P.~Toth and D.~Vigo, \emph{Vehicle routing: problems, methods, and
  applications}.\hskip 1em plus 0.5em minus 0.4em\relax SIAM, 2014.

\bibitem{3}
H.~N. Psaraftis, ``A dynamic programming solution to the single vehicle
  many-to-many immediate request dial-a-ride problem,'' \emph{Transportation
  Science}, vol.~14, no.~2, pp. 130--154, 1980.

\bibitem{4}
J.-F. Cordeau, ``A branch-and-cut algorithm for the dial-a-ride problem,''
  \emph{Operations Research}, vol.~54, no.~3, pp. 573--586, 2006.

\bibitem{2}
J.-F. Cordeau and G.~Laporte, ``A tabu search heuristic for the static
  multi-vehicle dial-a-ride problem,'' \emph{Transportation Research Part B:
  Methodological}, vol.~37, no.~6, pp. 579--594, 2003.

\bibitem{5}
S.~Ropke, J.-F. Cordeau, and G.~Laporte, ``Models and branch-and-cut algorithms
  for pickup and delivery problems with time windows,'' \emph{Networks},
  vol.~49, no.~4, pp. 258--272, 2007.

\bibitem{6}
K.~Braekers, A.~Caris, and G.~K. Janssens, ``Exact and meta-heuristic approach
  for a general heterogeneous dial-a-ride problem with multiple depots,''
  \emph{Transportation Research Part B: Methodological}, vol.~67, pp. 166--186,
  2014.

\bibitem{7}
C.~F. Daganzo, ``An approximate analytic model of many-to-many demand
  responsive transportation systems,'' \emph{Transportation Research}, vol.~12,
  no.~5, pp. 325--333, 1978.

\bibitem{8}
J.-J. Jaw, A.~R. Odoni, H.~N. Psaraftis, and N.~H. Wilson, ``A heuristic
  algorithm for the multi-vehicle advance request dial-a-ride problem with time
  windows,'' \emph{Transportation Research Part B: Methodological}, vol.~20,
  no.~3, pp. 243--257, 1986.

\bibitem{14}
J.~Guo and C.~Liu, ``A spatio-temporal based parallel insertion algorithm for
  solving dial-a-ride problem,'' in \emph{the 26th Chinese Control and Decision
  Conference (2014 CCDC), IEEE}, 2014, pp. 3257--3261.

\bibitem{9}
M.~Maalouf, C.~A. MacKenzie, S.~Radakrishnan, and M.~Court, ``A new fuzzy logic
  approach to capacitated dynamic dial-a-ride problem,'' \emph{Fuzzy Sets and
  Systems}, vol. 255, pp. 30--40, 2014.

\bibitem{11}
A.~Attanasio, J.-F. Cordeau, G.~Ghiani, and G.~Laporte, ``Parallel tabu search
  heuristics for the dynamic multi-vehicle dial-a-ride problem,''
  \emph{Parallel Computing}, vol.~30, no.~3, pp. 377--387, 2004.

\bibitem{12}
A.~Beaudry, G.~Laporte, T.~Melo, and S.~Nickel, ``Dynamic transportation of
  patients in hospitals,'' \emph{OR spectrum}, vol.~32, no.~1, pp. 77--107,
  2010.

\bibitem{13}
D.~Kirchler and R.~W. Calvo, ``A granular tabu search algorithm for the
  dial-a-ride problem,'' \emph{Transportation Research Part B: Methodological},
  vol.~56, pp. 120--135, 2013.

\bibitem{10}
F.~Glover, ``Future paths for integer programming and links to artificial
  intelligence,'' \emph{Computers \& operations research}, vol.~13, no.~5, pp.
  533--549, 1986.

\bibitem{16}
W.~P. Nanry and J.~W. Barnes, ``Solving the pickup and delivery problem with
  time windows using reactive tabu search,'' \emph{Transportation Research Part
  B: Methodological}, vol.~34, no.~2, pp. 107--121, 2000.

\bibitem{17}
S.~N. Parragh and V.~Schmid, ``Hybrid column generation and large neighborhood
  search for the dial-a-ride problem,'' \emph{Computers \& Operations
  Research}, vol.~40, no.~1, pp. 490--497, 2013.

\end{thebibliography}

\end{document}